\documentclass[11pt]{article}

\usepackage[final]{acl}

\usepackage{times}
\usepackage{latexsym}

\usepackage[T1]{fontenc}

\usepackage[utf8]{inputenc}

\usepackage{microtype}
\usepackage{fancyhdr}

\usepackage{inconsolata}

\usepackage{graphicx}

\usepackage{graphicx}
\usepackage{amsmath}
\usepackage{amsfonts}
\usepackage{algorithm}
\usepackage{tcolorbox}
\usepackage{algorithmic}
\usepackage{multirow} 
\usepackage{enumitem}
\usepackage{dashrule}
\usepackage{arydshln}
\usepackage{pifont}
\usepackage{booktabs}
\usepackage{array} 
\usepackage{subcaption}
\usepackage{placeins}
\usepackage{dblfloatfix}
\usepackage{hyperref}
\usepackage{url}
\usepackage{xcolor}
\tcbuselibrary{skins, breakable, listings, hooks}
\usepackage{tikz}
\usepackage{overpic}
\usetikzlibrary{arrows.meta, positioning, decorations.pathmorphing}
\definecolor{myblue}{RGB}{222,235,247} 
\definecolor{mydarkblue}{RGB}{0,176,240}
\definecolor{mydarkgreen}{RGB}{0,176,80}
\definecolor{mydarkyellow}{RGB}{255,192,0}
\definecolor{myorange}{RGB}{237,125,49}

\title{UR$^2$: Unify RAG and Reasoning through Reinforcement Learning}

\author{
\textbf{Weitao Li$^{1,2}$}, \textbf{Boran Xiang$^{3}$}, \textbf{Xiaolong Wang$^{1,2}$}, \textbf{Jingyi Ren$^{1,2}$}, \textbf{Ante Wang$^{2}$},\\
\textbf{Zhinan Gou$^{3}$}, \textbf{Weizhi Ma$^{2,\dagger}$}, \textbf{Yang Liu$^{1,2,\dagger}$}\\
\small $^1$ Dept. of Comp. Sci. \& Tech., Institute for AI, Tsinghua University, Beijing, China \\
\small $^2$ Institute for AI Industry Research (AIR), Tsinghua University, Beijing, China \\
\small $^3$ School of Management Science \& Information Engineering, Hebei University of Economics and Business, Hebei, China
}

\date{
    \small
    $^1$Dept. of Computer Science \& Technology, Institute for AI, Tsinghua University, Beijing, China \\
    $^2$Institute for AI Industry Research (AIR), Tsinghua University, Beijing, China \\
    $^3$School of Management Science and Information Engineering, Hebei University of Economics and Business, Hebei, China
}
\begin{document}

\thispagestyle{fancy}
\fancyhf{}
\rhead{Published as a conference paper at ACL 2026 (Oral Presentation)}

\maketitle
\begin{abstract}

Large Language Models (LLMs) have shown strong capabilities through two complementary paradigms: Retrieval-Augmented Generation (RAG) for knowledge grounding and Reinforcement Learning from Verifiable Rewards (RLVR) for complex reasoning. However, existing attempts to unify these paradigms remain narrow in scope, typically limited to open-domain QA with fixed retrieval settings, which constrains generalization to broader domains.
To address this limitation, we propose \textbf{UR$^2$} (\textbf{U}nified \textbf{R}AG and \textbf{R}easoning), a general reinforcement learning framework that dynamically coordinates retrieval and reasoning. UR$^2$ introduces two key designs: a difficulty-aware curriculum that selectively invokes retrieval only for challenging instances, and a hybrid knowledge access strategy that combines domain-specific offline corpora with on-the-fly LLM-generated summaries. Together, these components mitigate the imbalance between retrieval and reasoning and improve robustness to noisy information.
Experiments on open-domain QA, MMLU-Pro, medical, and mathematical reasoning tasks show that UR$^2$, built on Qwen-2.5-3/7B and LLaMA-3.1-8B, consistently outperforms existing RAG and RL baselines, and achieves performance comparable to GPT-4o-mini and GPT-4.1-mini on several benchmarks. Code, models, and data are available at \href{https://github.com/Tsinghua-dhy/UR2}{GitHub}.

\end{abstract}

{\let\thefootnote\relax\footnotetext{$\dagger$ Weizhi Ma and Yang Liu are corresponding authors.}}

\section{Introduction}

Large Language Models (LLMs) have achieved remarkable performance across diverse tasks by incorporating external knowledge (Retrieval-Augmented Generation, RAG)~\citep{lewis2020retrieval,borgeaud2022improving,izacard2022few} and optimizing reasoning through reinforcement learning with verifiable rewards (RLVR)~\citep{guo2025deepseek}. RAG methods enable LLMs to access external knowledge, while RLVR shows strong gains on mathematical and logical reasoning~\citep{zeng2025simplerl,chen2025empirical}. Motivated by these successes, recent work has begun to integrate retrieval and reasoning: for example, Search‑o1~\citep{li2025search} embeds an agentic RAG workflow into the LLM's chain‑of‑thought, and RAG‑Gym~\citep{xiong2025rag} proposes a unified RL‑based training framework for RAG agents. Similarly, RAG-RL methods which learn to invoke retrieval through RL---such as R1-Searcher~\citep{song2025r1} and Search-R1~\citep{jin2025search} use RLVR to train models on \emph{when} and \emph{what} to retrieve during reasoning, improving performance in open-domain QA.

Despite recent progress, RAG-RL frameworks remain limited in scope. Most methods focus narrowly on open-domain QA, with retrieval tied to fixed reasoning steps or static knowledge sources like Wikipedia. However, paradigms that work well on open-domain QA often fail to transfer to broader domains. \textbf{Two key limitations persist}: (1) \textbf{over-reliance on retrieval}, where models issue ill-posed or trivial queries to offload reasoning and weaken internal inference; (2) \textbf{collapse to pure reasoning}, where retrieval is avoided due to noisy documents and ineffective integration, causing the model to degenerate into standalone Chain-of-Thought reasoning. For instance, R1-Searcher and Search-R1 assume access to Wikipedia, ill-suited for tasks requiring specialized information. While methods like DeepResearcher attempt training in real web environments, they face inefficiencies due to the noisy and unstructured nature of online data~\citep{zheng2025deepresearcher}. Other methods like ZeroSearch~\citep{sun2025zerosearch}, use LLM-generated corpora to simulate retrieval, avoiding API costs but risking hallucination and loss of real-world complexity.

To address the limitations of existing RAG-RL approaches such as static retrieval, limited domain generalization, and poor robustness in noisy environments---we propose a general and adaptive framework, \textbf{UR$^2$} (\textbf{U}nified \textbf{R}AG and \textbf{R}easoning), which uses RL to dynamically coordinate retrieval and reasoning. 
Unlike prior methods that rely solely on static corpora (e.g., Wikipedia) or simulate retrieval with synthetic content, UR$^2$ combines both: it leverages task-specific offline corpora for accurate grounding, augmented with on-the-fly LLM-generated summaries for efficiency and generalization. 
To address the \textbf{imbalance} between retrieval and reasoning in prior methods, we design a difficulty-aware curriculum that adaptively controls when to trigger retrieval during training. Specifically, retrieval is used only for hard instances, encouraging the model to rely on internal reasoning when possible and to learn retrieval strategies only when necessary. This reduces retrieval overhead, improves query quality on challenging questions, and preserves reasoning capability across tasks.

We train UR$^2$ on Qwen-2.5-3B/7B-Instruct~\citep{yang2024qwen2} and LLaMA-3.1-8B-Instruct~\citep{dubey2024llama} across MMLU-Pro, Medicine, Math, and open-domain QA. During training, these models spontaneously develop key cognitive behaviors: self-verification through retrieval, intermediate reasoning validation, and hypothesis revision based on external evidence. UR$^2$ outperforms previous state-of-the-art (SOTA) methods by \textbf{5.8\%} (7B) and \textbf{19.0\%} (3B) on average, with peak gains of \textbf{9.5\%} and \textbf{29.6\%}. Notably, our 7B model matches GPT-4o-mini and GPT-4.1-mini\footnote{https://chat.openai.com/} , and generalizes well across domains and model architectures.

\paragraph{Our main contributions are summarized as follows:}
\begin{itemize}
\item We propose the first unified retrieval-reasoning RL framework that adapts to diverse tasks beyond open-domain QA, representing an important milestone for AI systems combining parametric and external knowledge.
\item We design a unified data representation and training scheme bridging retrieval and reasoning, with difficulty-aware curricula and and LLM-based summarization of retrieved evidence for accurate grounding and broad generalization, providing a solution to the imbalance between retrieval and reasoning.
\item Comprehensive experiments demonstrate that UR$^2$ surpasses advanced RAG and RL methods without expert demonstrations and generalizes robustly across domains.
\end{itemize}

\section{Related Work}

\subsection{Retrieval‑Augmented Generation}

RAG enhances LLMs by incorporating external information to reduce hallucinations~\citep{gao2023retrievalsurvey}. Early RAG methods concatenate retrieved documents with input prompts~\citep{lewis2020retrieval,izacard2022few,borgeaud2022improving}. Subsequent approaches have evolved in multiple directions: advanced RAG methods incorporate sophisticated retrieval and re-ranking strategies~\citep{gao2023retrievalsurvey,peng2024graph}; post-hoc verification methods address hallucinations by retrieving documents based on generated responses~\citep{li-etal-2024-citation,sun-etal-2024-towards-verifiable}; and Graph-based RAG methods integrate knowledge graphs for multihop reasoning~\citep{edge2024local, hu-etal-2025-grag,peng2024graph}. Recent RAG-RL frameworks have explored retrieval integration during training via real-time or synthetic retrieval~\citep{zheng2025deepresearcher,sun2025zerosearch}. However, these approaches remain constrained by static retrieval strategies, limited domain generalization, and \textbf{inability to dynamically coordinate retrieval with reasoning} across diverse task types.

\subsection{Reinforcement Learning for Retrieval‑Enhanced Reasoning}

RL has emerged as a key technique for significantly improving LLM capabilities, evolving from early policy gradient methods such as REINFORCE~\citep{williams1992simple} to more advanced algorithms like PPO~\citep{schulman2017proximal} and GRPO~\citep{shao2024deepseekmath}. 
Recent methods, including ARENA~\citep{ren2025effective}, Search‑R1~\citep{jin2025search}, and R1‑Searcher~\citep{song2025r1}, demonstrate that RL enables LLMs to effectively learn multi-step reasoning and retrieval strategies without requiring human feedback. 
These works collectively highlight a clear shift from fixed retrieval heuristics to learned, RL-driven retrieval policies, which form the foundation for our unified framework, with retrieval becoming \textbf{increasingly parameterized} rather than merely prompt-guided.

\section{Method}

We propose UR$^2$, a unified framework that integrates retrieval-based grounding with explicit step-by-step reasoning through reinforcement learning.

\begin{figure*}[ht!]
  \centering
  \includegraphics[width=\linewidth]{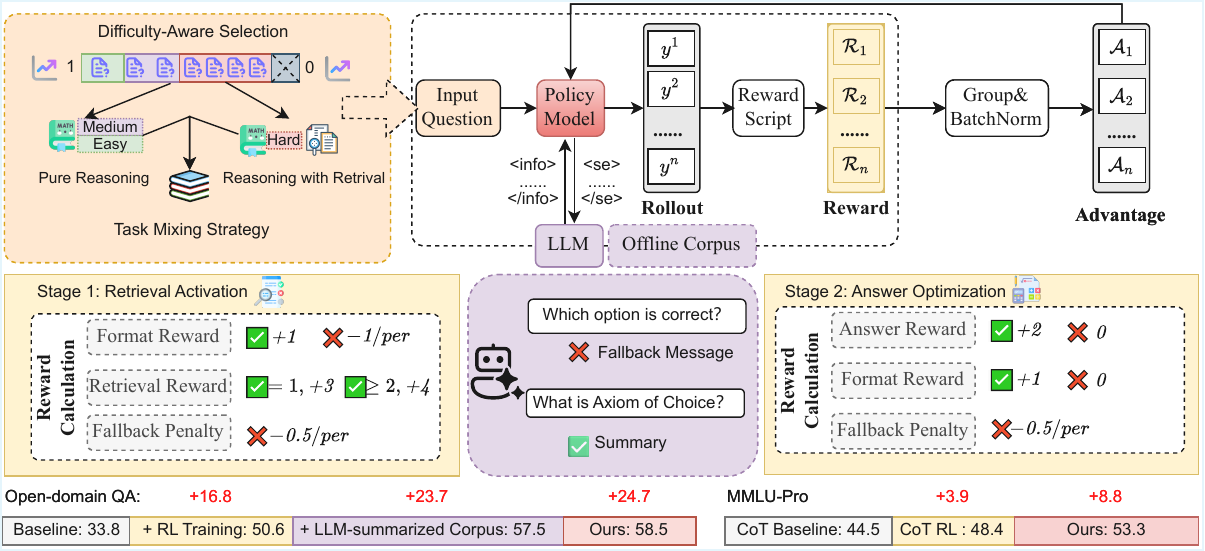}
  \caption{
    Overview of the UR$^2$ training pipeline. The top illustrates LLM-summarized retrieval corpus, difficulty-aware curriculum design and a two-stage reward design for retrieval activation and answer optimization. The bottom horizontal bars indicate: (1) Open-domain QA: ablation results under F1 score; and (2) MMLU-Pro: comparison with baselines using EM score.
  }
  \label{fig:ur2_method}
\end{figure*}

\noindent When extending retrieval-augmented generation to reasoning-intensive domains (e.g., mathematical problem solving and domain-specific QA), we observe a fundamental challenge: \emph{an imbalance between retrieval and reasoning}. Specifically, models may either over-rely on retrieval, offloading reasoning to external evidence, or collapse to pure chain-of-thought reasoning while failing to utilize external knowledge when necessary.

UR$^2$ is designed to address this imbalance in a principled and systematic manner.
Rather than treating retrieval as an always-on module or a purely auxiliary component, our framework explicitly coordinates \emph{when} to retrieve, \emph{how} retrieved information is presented, and \emph{which training signals encourage balanced behavior}.
To this end, UR$^2$ combines (i) on-the-fly LLM-based summarization that transforms retrieved evidence into compact, reasoning-compatible representations, and (ii) a difficulty-aware curriculum that selectively activates retrieval based on task hardness and knowledge demands.
Together, these components enable UR$^2$ to generalize across reasoning-intensive tasks while avoiding both retrieval overuse and reasoning degradation.

\subsection{Imbalance Between Retrieval and Reasoning}
\label{sec:imbalance}

When extending retrieval-augmented generation to reasoning-intensive tasks in our experiments, we observe a systematic imbalance between retrieval and reasoning behaviors: 1) \textbf{Over-reliance on retrieval.} In a controlled setting, we remove constraints on retrieval usage and allow the model to freely issue retrieval queries during RL training. We observe that the model frequently generates ill-posed or trivial queries (e.g., ``which option is correct''), effectively offloading reasoning to retrieval and weakening its internal reasoning process. This behavior leads to unstable training and degraded downstream performance, indicating that unconstrained retrieval is harmful in reasoning-centric tasks. 2) \textbf{Collapse to pure reasoning.} Conversely, under a Search-R1–style training setup, where retrieval summaries are removed and the model is trained with raw or noisy retrieved documents, training becomes unstable and retrieval is gradually abandoned. In this case, the model degenerates to pure chain-of-thought reasoning, failing to leverage external knowledge even for knowledge-intensive questions. This phenomenon highlights that retrieval noise directly interferes with reasoning and discourages effective retrieval usage.

These observations indicate that effective RAG for reasoning requires explicit mechanisms to regulate retrieval activation and its interaction with the reasoning process.

\definecolor{seColor}{RGB}{30,90,160}
\definecolor{infoColor}{RGB}{120,50,130}
\definecolor{goodGray}{RGB}{40,120,120}
\definecolor{warnOrange}{RGB}{180,60,30}

\newcommand{\se}[1]{\textcolor{seColor}{\texttt{$<$se$>$}}~#1~\textcolor{seColor}{\texttt{$<$/se$>$}}}
\newcommand{\info}[1]{\textcolor{infoColor}{\texttt{$<$info$>$}}~#1~\textcolor{infoColor}{\texttt{$<$/info$>$}}}

\begin{figure*}[h!]
\centering
\begin{tcolorbox}[
  colback=gray!4, colframe=gray!60!black, boxrule=0.6pt,
  fonttitle=\bfseries, fontupper=\small, rounded corners, width=\textwidth]
You are solving a multiple-choice question. Analyze each option step by step and select the best choice.
If you're uncertain about any fact, you may issue a \textbf{search query} like this: \se{a concise query (under 20 words)}
\begin{itemize}[left=1em, topsep=0.4em]
  \item You may issue \textbf{multiple queries} during your reasoning.
  \item Each query should focus on \textbf{only one specific fact or concept} and \textcolor{warnOrange}{\textbf{Avoid combining multiple facts in a single query.}}
  \item[] \textcolor{gray}{\textit{[Examples omitted here]}}
  \item You may use \textbf{up to four queries} in total --- use them wisely.
\end{itemize}
When documents are returned in the format: \info{... (search results here)}, integrate the retrieved information into your reasoning to refine your analysis and reach a well-supported conclusion.

Finally, give your answer in this format: \texttt{the correct answer is: A, B, C, D, etc.}
\end{tcolorbox}
\caption{Instruction prompt used to guide retrieval-augmented reasoning in UR$^2$. See Appendix~\ref{sec:eval_prompts} for details.}
\label{fig:ur2_prompt}
\end{figure*}

\subsection{Preventing Over-Reliance on Reasoning via LLM-Summarized Retrieval}
\label{sec:llm_corpus}

When retrieval is exposed to raw or noisy documents, models trained with RL tend to abandon retrieval altogether and collapse to pure chain-of-thought reasoning.
To prevent this failure mode, UR$^2$ introduces an LLM-summarized retrieval mechanism that converts retrieved content into compact and reasoning-compatible evidence.

UR$^2$ accesses domain-relevant knowledge sources, including domain-specific resources (e.g., curated medical references or encyclopedic content).
Instead of directly feeding retrieved documents to the model, retrieved content is transformed into concise, reasoning-compatible representations using an LLM, avoiding the introduction of misleading or low-quality evidence.

Beyond noise reduction, the summarization mechanism provides a substantial computational benefit. In online retrieval settings, retrieved documents average 11,346 tokens, which the summarizer compresses to 761 tokens---a \textbf{14.91$\times$} compression ratio (see Appendix~\ref{sec:compression_ratio}). Given the quadratic complexity of the attention mechanism, this compression trades a small sequential summarization cost for a much larger reduction in reasoning-time computation, often making the overall pipeline faster despite the additional summarization step. This design ensures that retrieval remains a usable and beneficial capability during optimization, preventing degeneration into pure reasoning even in knowledge-intensive or noisy environments.

\subsection{Preventing Over-Reliance on Retrieval via Difficulty-Aware Curriculum}
\label{sec:curriculum_design}

Conversely, unconstrained retrieval usage during training can cause models to offload reasoning to external search, leading to shallow reasoning behaviors and ill-posed retrieval queries.
To avoid this failure mode, UR$^2$ employs a difficulty-aware curriculum that explicitly regulates when retrieval is used during training.

\paragraph{Training Data Selection}
We categorize training instances by difficulty to control retrieval exposure.
For each question, we perform 20 rollouts using Qwen-2.5-7B-Instruct and compute the average performance score ($s$), following~\citep{song2025r1,huang2025rag,sun2025improving}.
Based on $s$, questions are grouped into Easy ($0.8 \leq s \leq 1.0$), Medium ($0.5 \leq s < 0.8$), and Hard ($0.2 \leq s < 0.5$) levels.
Following prior work~\citep{yu2025dapo, guo2025synthetic}, extremely difficult samples ($s < 0.2$) are filtered, as they hinder stable learning.
We adopt a sampling ratio of 7:2:1 for hard, medium, and easy questions, prioritizing challenging instances while retaining sufficient direct reasoning cases.

\paragraph{Task Mixing Strategy}
Based on difficulty, we selectively assign retrieval-augmented or pure reasoning training formats.
For mathematical reasoning tasks, only hard problems invoke retrieval-augmented prompting (Figure~\ref{fig:ur2_prompt}), while easy and medium problems rely on direct step-by-step reasoning.
In contrast, open-domain QA consistently uses retrieval due to its inherent knowledge dependency.

By associating retrieval usage with task difficulty, this curriculum prevents retrieval from becoming a default behavior.
The model learns to rely on external knowledge only when necessary, preserving intrinsic reasoning ability while still benefiting from retrieval on knowledge-intensive problems. More detailed experimental configurations are provided in the Section~\ref{sec:experimental_settings} and Appendix~\ref{sec:training_set_details}.

\subsection{Two-Stage Optimization for UR$^2$}

Given the limited tool invocation capabilities of base models, especially in reasoning-integrated scenarios, we design a two-stage optimization framework to systematically develop retrieval skills and reasoning proficiency. We train UR$^2$ using REINFORCE++~\citep{guo2025deepseek}, a streamlined variant of PPO. To prevent overfitting to retrieved content, we adopt retrieval masking~\citep{song2025r1,jin2025search}. Our implementation is based on the REINFORCE++-baseline provided by OpenRLHF~\citep{hu2024openrlhf}. See Appendix~\ref{sec:training_set_details} and Section~\ref{sec:implementation_details} for detailed implementation. 

\paragraph{RAG-based Rollout}

UR$^2$ enables the model to issue retrieval queries during reasoning rather than pre-retrieving all information upfront. As illustrated in Figure~\ref{fig:ur2_prompt}, the prompting mechanism enforces key principles: queries target single facts grounded in external knowledge, retrieval occurs when needed during the reasoning process, and strict format constraints using special tokens demarcate retrieval actions. 

This design allows the model to strategically leverage external knowledge by learning \emph{when} to retrieve and \emph{what} to query for purposeful and grounded reasoning.

\paragraph{Stage 1: Retrieval Capability Activation}
\label{sec:stage1}
We use UR$^2$ with Qwen-2.5-7B-Instruct on mathematical and open-domain QA tasks as an example. In Stage 1, the model trains on mathematical problems requiring retrieval calls in the specified format (Figure~\ref{fig:ur2_prompt}). The objective is not answer accuracy, but to enforce correct usage of the retrieval tool and promote retrieval-invoking behavior. This specialized training runs for only 10 steps. Further details on task setup and extensions to other models are provided in Appendix~\ref{sec:stage1_training_details}.

The total reward is:
\begin{align}
R_{i,\text{stage1}} = R_{i,\text{format}} + R_{i,\text{retrieval}} - P_{i,\text{fallback}}
\end{align}
where (1) \textbf{Format Reward}: $+1$ for fully compliant output; $-1$ per violation (e.g., malformed tags, overlength queries, missing retrieval, or illegal tokens); (2) \textbf{Retrieval Reward}: $+3$ for one valid query, $+4$ for two or more; (3) \textbf{Fallback Penalty}: $-0.5$ per fallback fault. 

This stage equips the model with retrieval capabilities and promotes effective integration of retrieved information during generation.

\paragraph{Stage 2: Answer Quality Optimization}

Building on Stage 1, we incorporate correctness feedback to refine generation quality while preserving retrieval behaviors. The updated reward function is:
\begin{align}
R_{\text{i, stage2}} = R_{i, \text{answer}} + R_{i, \text{format}} - P_{i, \text{fallback}}
\end{align}
where (1) \textbf{Answer Reward}: $+2$ for correct answers, $0$ for incorrect; (2) \textbf{Format Reward}: $+1$ for fully valid format; $0$ otherwise; (3) \textbf{Fallback Penalty}: $-0.5$ per fallback fault.

By decoupling retrieval skill acquisition (Stage 1) from generation optimization (Stage 2), we ensure stable convergence and interpretable credit assignment across complex reasoning trajectories.

\section{Experimental Settings}
\label{sec:experimental_settings}

\subsection{Training Datasets}

We build a unified training set covering math (SimpleZoo-RL~\citep{zeng2025simplerl}), open-domain QA (R1-Searcher~\citep{song2025r1}), and multi-choice medical QA (MedQA~\citep{jin2021disease}). For MMLU-Pro\citep{wang2024mmlu} domains (philosophy, history, economics), we generate synthetic questions via \texttt{Qwen-3-32B}\footnote{https://huggingface.co/Qwen/Qwen3-32B}. After deduplication and data selection using Qwen-2.5-7B-Instruct on 20 rollouts per question, we obtain 3K samples each for math, open-domain QA, and MedQA, and 2K samples for each MMLU-Pro domain. Details are in Appendix~\ref{sec:training_set_details}.

\subsection{Evaluation Benchmarks}

We evaluate generalization across four task families: 
(1) \textbf{Math Reasoning}: MATH500 (in-domain)~\citep{hendrycks2021measuring}, Minerva (OOD)~\citep{lewkowycz2022solving}; metric: \texttt{LLM-as-a-judge}. 
(2) \textbf{Medical QA}: MedQA (5-choice, in)~\citep{jin2021disease}, MMLU-Pro Medical (M-Med, OOD); metric: EM. 
(3) \textbf{MMLU-Pro}: Philosophy, History, Economics (in), Law (OOD); metric: EM. 
(4) \textbf{Open-Domain QA}: HotpotQA~\citep{yang2018hotpotqa}, 2WikiMultiHopQA~\citep{ho-etal-2020-constructing} (in); Bamboogle~\citep{press2023measuring}, MusiQue~\citep{trivedi-etal-2022-musique} (OOD); metrics: F1 and \texttt{LLM-as-a-judge}.

\subsection{Baselines}

We compare UR$^2$ to: 
(1) \textbf{Vanilla Methods}: Chain-of-Thought~\citep{kojima2022large}, Standard RAG~\citep{borgeaud2022improving,izacard2022few} (top-$k$=10). 
(2) \textbf{Advanced RAG Methods}: Search-o1~\citep{li2025search}, Self-Ask~\citep{press2023measuring}, and RAT~\citep{wang2024rat}, which combine reasoning with retrieval using prompt. 
(3) \textbf{CoT-RL Methods}: R1-like methods including Open-Reasoner-Zero~\citep{hu2025open}, SimpleRL-Zoo~\citep{zeng2025simplerl}, and

\begin{table*}[h!]
\centering
\small

\setlength{\tabcolsep}{1.4pt} 
\begin{tabular}{l ccccc c@{\hspace{0.2em}}cc c@{\hspace{0.2em}}cc}
\toprule
\multirow{2}{*}{\textbf{Method}} 
& \multicolumn{5}{c}{\textbf{MMLU-Pro}} 
& \multicolumn{3}{c}{\textbf{Medicine}} 
& \multicolumn{3}{c}{\textbf{Math}} \\
\cmidrule(lr){2-6} \cmidrule(lr){7-9} \cmidrule(lr){10-12}
& Hist.$^\dag$ & Phil.$^\dag$ & Econ.$^\dag$ & Law$^\ddag$ & \textbf{Avg} 
& MedQA$^\dag$ & M-Med$^\ddag$ & \textbf{Avg} 
& Math500$^\dag$ & Minerva$^\ddag$ & \textbf{Avg} \\
\midrule
\multicolumn{12}{c}{\textbf{\textit{GPT-4o-mini}}} \\
CoT &56.7 &\underline{53.1} &\underline{70.4} &\textbf{38.2} &\underline{54.5} &71.4 &67.0 &69.2 &\underline{78.0} &65.6 &71.8 \\
Standard RAG &\underline{57.0} &52.3 &68.6 &\underline{36.2} &53.5 &70.6 &64.2 &67.4 &77.1 &\textbf{68.4} &\underline{72.8}  \\
\multicolumn{12}{c}{\fontsize{7.5pt}{9pt}\selectfont\textcolor{gray}{\textit{Advanced RAG Methods}}} \\
Self-Ask &56.3 &48.5 &67.8 &31.2 &51.0 &72.4 &\underline{68.0} &70.2 &62.9 &45.2 &54.1 \\
RAT &\textbf{57.5} &\textbf{55.3} &\textbf{73.0} &34.2 &\textbf{55.0} &\underline{74.4} &\textbf{70.6} &\textbf{72.5} &77.5 &64.2 &70.9  \\
Search-o1 &53.5 &\textbf{55.3} &69.8 &35.4 &53.5 &\textbf{75.2} &66.6 &\underline{70.9} &\textbf{78.6} &\underline{68.3} &\textbf{73.5}  \\
\midrule
\multicolumn{12}{c}{\textbf{\textit{Qwen-2.5-7B}}} \\
CoT  &42.3 &45.7 &63.4 &26.6 &44.5 &57.2 &52.0 &54.6 &76.6 &59.4 &68.0   \\
Standard RAG &44.6 &41.1 &57.8 &26.0 &42.4 &54.2 &53.2 &53.7 &73.8 &54.6 &64.2  \\
\multicolumn{12}{c}{\fontsize{8.0pt}{9.5pt}\selectfont\textcolor{gray}{\textit{CoT-RL Methods}}} \\
General Reasoner &47.9 &44.2 &65.9 &30.4 &47.1 &58.4 &54.4 &56.4 &76.6 &\textbf{62.1} &68.8  \\
Open-Reasoner-Zero &50.0 &46.6 &67.5 &\underline{34.2} &49.6 &61.6 &58.8 &60.2 &\underline{80.7} &58.8 &\underline{69.8}  \\
SimpleRL-Zoo &35.7 &36.9 &55.2 &25.4 &38.3 &57.2 &51.0 &54.1 &77.1 &50.7 &63.9  \\
\multicolumn{12}{c}{\fontsize{8.0pt}{9.5pt}\selectfont\textcolor{gray}{\textit{Our Implementations}}} \\
Vanilla RL &\underline{52.2} &43.5 &64.0 &33.8 &48.4 &64.2 &57.4 &60.8 &78.2 &59.4 &68.8  \\
\textbf{UR$^2$ (qw Summary)} &51.2 &\underline{49.7} &\underline{67.8} &32.6 &\underline{50.3} &\underline{67.2} &\underline{60.8} &\underline{64.0} &80.2 &59.4 &\underline{69.8}  \\
\textbf{UR$^2$ (GPT Summary)} &\textbf{53.2}& \textbf{53.0}& \textbf{72.2}& \textbf{35.0}& \textbf{53.3}& \textbf{69.6}& \textbf{62.8}& \textbf{66.2}& \textbf{80.9}& \underline{61.0} & \textbf{71.0}  \\
\midrule

\multicolumn{12}{c}{\textbf{{\textit{Qwen-2.5-3B}}}} \\
CoT &33.6 &32.3 &48.8 &20.6 &33.8 &39.4 &36.8 &38.1 &63.6 &39.9 &51.8  \\
Standard RAG &37.8 &36.5 &51.4 &23.2 &37.1 &45.6 &40.0 &42.8 &65.3 &40.8 &53.1  \\
\multicolumn{12}{c}{\fontsize{8.0pt}{9.5pt}\selectfont\textcolor{gray}{\textit{Our Implementations}}} \\
Vanilla RL &40.7 &34.7 &55.0 &24.6 &38.7 &51.8 &47.6 &49.7 &\underline{68.0} &43.9 &56.0 \\
\textbf{UR$^2$ (qw summary)} &\underline{43.0} &\underline{42.6} &\underline{60.6} &\underline{26.6} &\underline{43.2} &\underline{55.2} &\underline{54.0} &\underline{54.6} &\textbf{68.4} &\textbf{47.4} &\textbf{57.9}   \\
\textbf{UR$^2$ (GPT summary)} &\textbf{47.8} &\textbf{49.3} &\textbf{63.9} &\textbf{30.0} &\textbf{47.8} &\textbf{59.8} &\textbf{56.8} &\textbf{58.3} &67.8 &\underline{45.0} &\underline{57.2}   \\
\midrule

\multicolumn{12}{c}{\textbf{{\textit{LLaMA-3.1-8B}}}} \\
CoT &37.8 &\underline{40.9} &53.4 &\textbf{29.0} &40.3 &59.6 &52.6 &56.1 &48.4 &34.4 &41.4  \\
Standard RAG &43.6 &33.9 &51.0 &26.6 &38.8 &56.4 &53.2 &54.8 &45.0 &31.4 &38.2  \\
\multicolumn{12}{c}{\fontsize{8.0pt}{9.5pt}\selectfont\textcolor{gray}{\textit{Our Implementations}}} \\
Vanilla RL &\underline{44.6} &36.9 &53.0 &26.4 &40.2 &\underline{66.8} &\underline{57.4} &\underline{62.1} &45.5 &\textbf{43.4} &\underline{44.4}  \\

\textbf{UR$^2$ (qw summary)} &43.6 &\textbf{41.2} &\underline{56.8} &28.0 &\underline{42.4} &65.8 &56.2 &61.0 &\underline{48.6} &38.4 &43.5  \\
\textbf{UR$^2$ (GPT summary)} &\textbf{48.3} &38.6 &\textbf{58.0} &\underline{28.8} &\textbf{43.4} &\textbf{68.6} &\textbf{58.4} &\textbf{63.5} &\textbf{54.5} &\underline{39.0} &\textbf{46.8}  \\

\bottomrule
\end{tabular}
\caption{
Performance on reasoning and math tasks. 
We report EM scores (in \%) on MMLU-Pro and MedQA, and \texttt{LLM-as-a-judge} scores (in \%) on math benchmarks.
$\dag$ = in-domain, $\ddag$ = out-of-domain. 
Best results are \textbf{bold}; second-best are \underline{underlined}.
}
\label{tab:main_reasoning}

\end{table*}

\noindent General-Reasoner~\citep{ma2025general}. 
(4) \textbf{RAG-RL Methods}: R1-Searcher~\citep{song2025r1}, R1-Searcher++~\citep{song2025r1++}, Search-R1~\citep{jin2025search}, and ZeroSearch~\citep{sun2025zerosearch}. 
(5) \textbf{Vanilla RL}: Baseline implementation following the same training setup and datasets as UR$^2$, with RAG-RL applied to open-domain QA and CoT-RL to mathematical and multiple-choice tasks. Details are in Appendix~\ref{sec:vanilla_rl}.

We use Qwen-2.5-3B-Instruct, Qwen-2.5-7B-Instruct, LLaMA-3.1-8B-Instruct, GPT-4o-mini, and GPT-4.1-mini as backbones (see Appendix~\ref{sec:evaluation_details} for configs).

\subsection{Implementation Details}
\label{sec:implementation_details}

Retrieval uses \texttt{BGE-large-en-v1.5\footnote{https://huggingface.co/BAAI/bge-large-en-v1.5}} and the KILT~\citep{petroni2021kilt} Wikipedia corpus (100-word segments, 29M documents) following~\citep{song2025r1}. Open-domain QA uses Wikipedia abstract corpus\footnote{https://nlp.stanford.edu/projects/hotpotqa/enwiki-20171001-pages-meta-current-withlinks-abstracts.tar.bz2}. Unless otherwise noted, all models, \textbf{including baselines}, use \texttt{GPT-4.1-mini} as the summarizer during training and \texttt{GPT-4.1} during evaluation with top-$k$ = 10, while mathematical tasks are summarized by \texttt{Qwen-3-32B}. For evaluation, we sample 500 instances from each benchmark. We use $G=16$ rollouts. For fair comparison, we train a \texttt{Qwen-2.5-7B-Instruct} summarizer for both training and evaluation on reasoning tasks. Open-domain QA tasks are excluded, as summarization is already an integral component of both our method and the baselines. 7B and 8B models use training batch size 256, rollout batch size 64; 3B doubles both. Learning rate = 2e-6. Up to \textbf{4} retrieval turns are allowed. All models are trained for up to 2 epochs on 8$\times$A100 GPUs (${\sim}$20 hours / 160 GPU-hours). See Appendix~\ref{sec:training_set_details} and~\ref{sec:evaluation_details} for details.
\begin{table*}[h!]
\centering
\small
\setlength{\tabcolsep}{1.7pt}
\begin{tabular}{p{1.9cm}>{\centering\arraybackslash}p{2.3cm} lcccccccccc}
\toprule 
\multirow{2}{*}{\textbf{Models}} & \multirow{2}{*}{\textbf{Types}} & \multirow{2}{*}{\textbf{Methods}} & \multicolumn{2}{c}{\textbf{Hotpot$^\dag$}} & \multicolumn{2}{c}{\textbf{2Wiki$^\dag$}} & \multicolumn{2}{c}{\textbf{Bamb.$^\ddag$}} & \multicolumn{2}{c}{\textbf{MusiQ.$^\ddag$}} & \multicolumn{2}{c}{\textbf{Avg}} \\
\cmidrule(lr){4-5} \cmidrule(lr){6-7} \cmidrule(lr){8-9} \cmidrule(lr){10-11} \cmidrule(lr){12-13}
& & & F1 & LSJ & F1 & LSJ & F1 & LSJ & F1 & LSJ & F1 & LSJ \\
\midrule
\multirow{5}{*}{\textbf{GPT-4.1-mini}} & \multirow{2}{*}{Vanilla Methods} & CoT &43.7 &59.2 &\underline{48.6} &\textbf{60.8} &59.2 &\textbf{76.0} &28.3 &\textbf{35.4} &45.0 &57.9\\
&& Standard RAG & 54.5 & \underline{74.4} & 41.3 & 52.4 & 46.4 & 51.2 & 21.9 & 28.4 & 41.0 & 51.6 \\
\cmidrule{3-3}
&\multirow{3}{*}{Advanced RAG} & Self-Ask &\textbf{65.4} &\textbf{75.0} &\textbf{52.7} &57.4 &\textbf{71.7} &\underline{75.2} &\textbf{31.6} &\underline{35.0} &\textbf{55.4} &\textbf{60.7} \\
&& RAT &\underline{56.9} &64.2 &45.7 &49.0 &60.3 &62.4 &\underline{29.0} &31.4 &\underline{48.0} &51.8\\
&& Search-o1 &53.1 &74.0 &44.4 &\underline{60.6} &\underline{63.7} &71.2 &28.6 &33.4 &47.5 &\underline{59.8} \\
\midrule
\multirow{11}{*}{\textbf{Qwen-2.5-7B}} & \multirow{2}{*}{Vanilla Methods} & CoT &24.9 &31.0 &25.1 &27.6 &41.3 &43.2 &14.8 &12.2 &26.5 &28.5 \\
&& Standard RAG &49.2 &62.8 &32.8 &37.6 &38.9 &40.0 &14.4 &14.6 &33.8 &38.8 \\
\cmidrule{3-3}
&\multirow{4}{*}{RAG-RL} & R1-Searcher &\underline{71.8} &78.0 &57.9 &63.6 &56.5 &53.6 &33.3 &32.6 &54.8 &57.0\\
&& Search-R1 &\textbf{72.4} &\underline{78.8} &61.0 &63.8 &58.9 &56.8 &32.2 &32.0 &56.1 &57.9\\
&& R1-Searcher++ &59.0 &64.2 &\underline{61.2} &\underline{64.4} &60.8 &59.2 &33.8 &32.8 &53.7 &55.2\\
&& ZeroSearch &46.0 &50.4 &38.4 &38.6 &35.8 &38.4 &14.7 &13.8 &33.7 &35.3 \\
\cmidrule{3-3}
&\multirow{2}{*}{\begin{minipage}[c]{2.5cm}\centering Our Implementations\end{minipage}} & Vanilla RL &70.9 &\underline{78.8} &\underline{61.2} &62.4 &\underline{63.3} &\textbf{63.2} &\underline{34.4} &\underline{34.4} &\underline{57.5} &\underline{59.6} \\
 && \textbf{UR$^2$} &71.2 &\textbf{79.4} &\textbf{62.6} &\textbf{65.2} &\textbf{64.5} &\underline{62.4} &\textbf{35.8} &\textbf{34.6} &\textbf{58.5} &\textbf{60.4} \\
\midrule
\multirow{6}{*}{\textbf{Qwen-2.5-3B}} & \multirow{2}{*}{Vanilla Methods} & CoT &26.6& 27.2&22.7 &22.6 &31.2 &33.6 &11.3 &9.6 &23.0 &23.3 \\
&& Standard RAG &50.6 &57.0 &29.8 &30.4 &26.1 &27.2 &9.7 &7.4 &29.1 &30.5\\
\cmidrule{3-3}
&\multirow{2}{*}{RAG-RL} & Search-R1 &63.1 &69.2 &49.5 &53.4 &48.3 &48.0 &27.6 &27.8 &47.1 &49.6\\
&& Zero-Search &42.7 &45.8 &26.1 &27.6 &32.4 &31.2 &16.9 &17.0 &29.5 &30.4\\
\cmidrule{3-3}
&\multirow{2}{*}{\begin{minipage}[c]{2.5cm}\centering Our Implementations\end{minipage}} & Vanilla RL &\underline{65.9} &\underline{73.6} &\underline{54.9} &\underline{58.0} &\underline{59} &\underline{57.6} &\underline{30.0} &\underline{29.6} &\underline{52.5} &\underline{54.7} \\
&& \textbf{UR$^2$} &\textbf{67.7} &\textbf{76.0} &\textbf{55.2} &\textbf{58.6} &\textbf{57.8} &\textbf{58.4} &\textbf{30.5} &\textbf{31.6} &\textbf{55.3} &\textbf{56.2}
\\
\midrule
\multirow{5}{*}{\textbf{LLaMA-3.1-8B}} & \multirow{2}{*}{Vanilla Methods} & CoT &28.6 &31.6 &16.4&17.8 &43.0 &42.4 &9.8 &10.8 &24.5 &25.7 \\
&& Standard RAG &47.5 &54.4 &26.2 &26.4 &26.5 &28.0 &10.1 &10.2 &27.6 &29.8\\
\cmidrule{3-3}
&\multirow{1}{*}{RAG-RL} & R1-Searcher &\textbf{70.8} &76.8 &59.6 &62.2 &\textbf{64.7} &62.4 &31.1 &29.4 &\textbf{56.6} &57.7\\
\cmidrule{3-3}
&\multirow{2}{*}{\begin{minipage}[c]{2.5cm}\centering Our Implementations\end{minipage}} & Vanilla RL &70.0 &\underline{77.6} &\textbf{61.2} &\textbf{64.2} &60.6 &\textbf{63.2} &\underline{32.7} &\underline{31.8} &56.1 &\underline{59.2} \\
& & \textbf{UR$^2$} &\underline{70.1} &\textbf{78.8} &\underline{60.1} &\underline{63.2} &\underline{60.7}&\textbf{63.2} &\textbf{34.3} &\textbf{34.0} &\underline{56.3} &\textbf{59.8} \\
\bottomrule
\end{tabular}
\caption{Performance on open-domain QA tasks. We report F1 and \texttt{LLM-as-a-judge} (LSJ) scores, both in \%. $\dag$ = in-domain; $\ddag$ = out-of-domain.}
\label{tab:open_domain_qa}
\end{table*}
\section{Experimental Results}

Our UR$^2$ framework achieves SoTA performance across reasoning and retrieval tasks, enabling 7B models to match or exceed the GPT model family while significantly outperforming existing RAG and RL-based methods. More comprehensive baseline results can be found in Appendix~\ref{sec:comprehensive_baselines}.

\subsection{Main Results on Reasoning Tasks}

As shown in Table~\ref{tab:main_reasoning}, UR$^2$ (GPT Summary) demonstrates substantial improvements across all reasoning tasks on the Qwen-2.5-7B model, achieving average scores of 53.3\% on MMLU-Pro, 65.9\% on Medicine, and 71.0\% on Math benchmarks, representing gains of 3.7\%, 5.7\%, and 1.2\% over the strongest CoT-RL baseline Open-Reasoner-Zero. Using a distilled Qwen-2.5-7B summarizer instead of GPT-4.1 leads to a moderate performance drop, but UR$^2$ (qw Summary) consistently outperforms all non-UR$^2$ baselines, demonstrating that our gains stem from \textbf{the method itself rather than reliance on a strong proprietary summarization model}. Across model scales, UR$^2$ shows consistent advantages: on Qwen-2.5-3B, it achieves even larger performance gains with 9.1\% improvement on MMLU-Pro and 8.6\% on Medicine over Vanilla RL, demonstrating that UR$^2$ provides greater benefits for models with limited knowledge but strong reasoning capabilities. On LLaMA-3.1-8B, it achieves 43.4\% on MMLU-Pro, outperforming all baselines. Notably, our method achieves competitive performance with the more capable closed-source GPT-4o-mini model on several tasks. As shown in Tables~\ref{tab:main_reasoning} and~\ref{tab:reasoning_extra}, advanced RAG methods degrade performance on smaller models and require unacceptable source consumption (except Search-o1).
\FloatBarrier

\subsection{Main Results on Open-domain QA}

As demonstrated in Table~\ref{tab:open_domain_qa}, UR$^2$ achieves strong performance on open-domain QA, with Qwen-2.5-7B reaching 58.5\% average F1 score, outperforming the strongest RAG-RL baseline Search-R1 (56.1\%) by 2.4\%. UR$^2$ demonstrates particularly robust out-of-domain generalization, achieving 64.5\% on Bamboogle and 35.8\% on MusiQue, surpassing all baselines. Across model scales, UR$^2$ maintains consistent advantages: on Qwen-2.5-3B, 

\begin{table*}[h!]
\centering
\small
\begin{tabular}{l ccccc c@{\hspace{0.10em}}cc}
\toprule
\multirow{2}{*}{\textbf{Method}} 
& \multicolumn{5}{c}{\textbf{MMLU-Pro}} 
& \multicolumn{3}{c}{\textbf{Medicine}} \\
\cmidrule(lr){2-6} \cmidrule(lr){7-9}
& Hist.$^\dag$ & Phil.$^\dag$ & Econ.$^\dag$ & Law$^\ddag$  & \textbf{Avg}
& MedQA$^\dag$ & M-Med$^\ddag$  & \textbf{Avg} \\
\midrule
\textbf{UR$^2$} & \textbf{53.2} & \textbf{53.1} & \textbf{72.2} & 35.0 &\textbf{53.3} & 69.6 & 62.8 &66.2 \\
w/o Stage-1 & 48.0 & 51.1 & 68.0 & 30.9 &49.5 & 67.6 & 63.0 &65.3 \\
w/o $P_{\text{fallback}}$ & 52.0   & 51.3   & 68.4   & \textbf{36.6} &52.1   & \textbf{71.4} &62.0 &66.7 \\
w/o Task Mixing& 52.2  &  51.9  & 68.2   & 33.2  &51.4  & 70.0 & \textbf{63.6} &\textbf{66.8}  \\
w/o LLM Summary & --   & --   & --  & -- & --  & --   & --   &  -- \\
\midrule
Vanilla RL &52.2& 43.5& 64.0& 33.8 &48.4 &64.2 &57.4 &60.8\\
4omini Summary & 49.3  & 48.8   & 67.4   & 32.4 &49.5   & 65.0   & 59.2 &62.1   \\
Qw3-8B  Summary &  49.1  & 49.9   & 67.8   & 30.6 &49.4  & 64.8   & 58.2 &61.5  \\
Distilled Qw2.5-7B Summary &  49.6  & 52.3   & 68.2   & 31.4 &50.4  & 67.2   & 61.8 &64.5  \\
\bottomrule
\end{tabular}
\caption{Ablation study of Qwen-2.5-7B-Instruct on MMLU-Pro and medical reasoning tasks. ``w/o Task Mixing'' means retrieving for all samples. $\dag$ = in-domain; $\ddag$ = out-of-domain.}
\label{tab:mcqablation}
\end{table*}
\noindent it achieves 55.3\% F1, improving 8.2\% over Search-R1; on LLaMA-3.1-8B, it reaches 56.3\%, competitive with specialized RAG-RL methods. Notably, our 7B model surpasses GPT-4.1-mini (55.4\%) by 3.1\%, demonstrating UR$^2$'s effective dynamic coordination of retrieval and reasoning.
\vspace{-3pt}
\section{Analysis}
\label{sec:ablation}
\vspace{-3pt}
\textbf{Additional results are provided in the Appendix}, including further ablation studies (Appendix~\ref{sec:additional_ablation}), the impact of LLM summaries and corpus design (Appendix~\ref{sec:corpus_ablation}), comparative analysis of retrieval integration in RL training (Appendix~\ref{sec:training_analysis}), analysis of retrieval behavior (Appendix~\ref{sec:retrieval_behavior}), unsuccessful attempts on reasoning models (Appendix~\ref{sec:unsuccessful_attempts}), the compression ratio of the summarization module (Appendix~\ref{sec:compression_ratio}), training efficiency and latency analysis (Appendix~\ref{sec:efficiency}), and illustrative case studies (Appendix~\ref{case_study}).
\vspace{-3pt}
\subsection{Ablation Study}

\paragraph{Effect of the Summarization Model.}
We first examine the role of the summarization model used during training. As shown in Table~\ref{tab:mcqablation}, removing LLM-based summarization (\textit{w/o LLM Summary}) leads to complete training failure, where the model collapses into pure chain-of-thought reasoning and abandons retrieval entirely. This highlights that \textbf{retrieval noise is a fundamental challenge in RAG-RL training}, and that effective summarization is crucial for stabilizing retrieval usage. Replacing the default summarizer with alternative or weaker models (4omini, Qw3-8B) results in consistent 3--4\% performance drops across tasks, but still substantially outperforms vanilla RL. Notably, the \textit{Distilled Qw2.5-7B Summary} variant exhibits only moderate degradation, demonstrating that UR$^2$ does not rely on a strong proprietary summarization model and remains robust under constrained summarization capacity.

\paragraph{Ablation of Training Components.}
We further analyze ablations on key training components. The \textit{w/o Stage-1} variant shows notable performance drops (e.g., 5.2\% in History and 4.2\% in Economics), confirming that explicit retrieval activation is essential for learning effective retrieval behavior. Removing the fallback penalty (\textit{w/o $P_{\text{fallback}}$}) slightly improves performance on Law and MedQA, but frequently produces unreasonable or ill-posed queries such as ``\textbf{which option is right}'', indicating the necessity of regulating fallback behavior. The \textit{w/o Task Mixing} variant yields only minor performance changes, suggesting that selective retrieval primarily improves computational efficiency while maintaining or slightly enhancing accuracy. Overall, these results show that the two-stage optimization and carefully designed reward components jointly contribute to stable and effective retrieval-augmented reasoning.
\vspace{-5pt}
\subsection{Preservation of Intrinsic Reasoning Ability}
\begin{table}[h!]
\centering
\small
\begin{tabular}{lcccc}
\toprule
Method & MedQA & M-Med & Math500 & Minerva \\
\midrule
CoT & 57.2 & 52.0 & 76.6 & 59.4 \\
Vanilla-RL & 64.2 & 57.4 & 78.2 & 59.4 \\
UR$^2$-CoT & 59.8 & 58.8 & 77.1 & 61.9 \\
UR$^2$ & \textbf{69.6} & \textbf{62.8} & \textbf{80.9} & \textbf{61.0} \\
\bottomrule
\end{tabular}
\caption{Intrinsic reasoning under direct prompts (no retrieval). UR$^2$ preserves reasoning ability while retrieval further boosts performance.}
\label{tab:reasoning_preservation}
\end{table}
A key concern is whether retrieval-augmented training degrades standalone reasoning. We evaluate UR$^2$ (Qwen-2.5-7B) under direct reasoning prompts identical to Vanilla-RL, without retrieval. As shown in Table~\ref{tab:reasoning_preservation}, UR$^2$-CoT outperforms CoT and matches Vanilla-RL across knowledge-intensive and reasoning-intensive benchmarks, confirming that difficulty-aware curriculum learning preserves and strengthens intrinsic reasoning.

\begin{table*}[t!]
\centering
\small
\setlength{\tabcolsep}{1.3pt}
\begin{subtable}[t]{\textwidth}
\centering
\begin{tabular}{l ccccc ccc ccc}
\toprule
\multirow{2}{*}{\textbf{Corpus}} 
& \multicolumn{5}{c}{\textbf{MMLU-Pro}} 
& \multicolumn{3}{c}{\textbf{Medicine}} 
& \multicolumn{3}{c}{\textbf{Math}} \\
\cmidrule(lr){2-6} \cmidrule(lr){7-9} \cmidrule(lr){10-12}
& Hist.$^\dag$ & Phil.$^\dag$ & Econ.$^\dag$ & Law$^\ddag$ & \textbf{Avg} 
& MedQA$^\dag$ & M-Med$^\ddag$ & \textbf{Avg} 
& Math500$^\dag$ & Minerva$^\ddag$ & \textbf{Avg} \\
\midrule
\textbf{Local Corpus} 
& 53.2 & 53.0 & 72.2 & \textbf{35.0} & 53.3 
& 69.6 & 62.8 & 65.9 
& \textbf{80.9} & 61.0 & \textbf{71.0} \\
\textbf{Online Search} 
& \textbf{57.7} & \textbf{57.8} & 71.0 & \textbf{35.0} & \textbf{55.4} 
& \textbf{70.4} & \textbf{65.4} & \textbf{67.9}
& 78.7 & \textbf{61.2} & 70.0 \\
\bottomrule
\end{tabular}
\end{subtable}

\vspace{-1.8pt}

\setlength{\tabcolsep}{6.5pt}
\begin{subtable}[t]{\textwidth}
\centering
\begin{tabular}{l cc cc cc cc cc}
\toprule
\multirow{2}{*}{\textbf{Corpus}} 
& \multicolumn{2}{c}{\textbf{Hotpot$^\dag$}} 
& \multicolumn{2}{c}{\textbf{2Wiki$^\dag$}} 
& \multicolumn{2}{c}{\textbf{Bamb.$^\ddag$}} 
& \multicolumn{2}{c}{\textbf{MusiQ.$^\ddag$}} 
& \multicolumn{2}{c}{\textbf{Avg}} \\
\cmidrule(lr){2-3} \cmidrule(lr){4-5} \cmidrule(lr){6-7} \cmidrule(lr){8-9} \cmidrule(lr){10-11}
& F1 & LSJ & F1 & LSJ & F1 & LSJ & F1 & LSJ & F1 & LSJ \\
\midrule
\textbf{Local Corpus} 
& \textbf{71.2} & \textbf{79.4} 
& 62.6 & 65.0
& 64.5 & 62.4
& \textbf{35.8} & 34.6 
& 58.5 & 60.4 \\
\textbf{Online Search} 
& 62.0 & 67.6 
& \textbf{75.8} & \textbf{81.8} 
& \textbf{73.7} & \textbf{76.0} 
& 34.9 & \textbf{37.8} 
& \textbf{61.6} & \textbf{65.8} \\
\bottomrule
\end{tabular}

\end{subtable}
\caption{
Comparison of UR$^2$ Qwen-2.5-7B-Instruct using Local Corpus vs. Online Search across different tasks. $\dag$ = in-domain; $\ddag$ = out-of-domain.
}
\label{tab:online_search}
\end{table*}

\subsection{Impact of Online Search}

To test scalability under online retrieval, we compare local corpus with real-time search (Table~\ref{tab:online_search}). Online search yields consistent gains on MMLU-Pro (+2.1\% average) and medical tasks (+2.0\% average), and substantial improvements on 2Wiki (+13.2\% F1) and Bamboogle (+9.2\% F1), demonstrating strong generalization to scenarios requiring up-to-date or non-Wikipedia knowledge. The only exceptions are math---where \texttt{Qwen-3-32B}'s parametric knowledge already covers the required formulas---and HotpotQA, where rate limits block access to many gold Wikipedia pages. Notably, our setting does not enforce full top-$10$ coverage, which introduces noise and better reflects real-world retrieval conditions. The substantial gains on out-of-domain benchmarks (Bamboogle, MusiQue) are particularly noteworthy, as they demonstrate that UR$^2$'s learned retrieval strategies transfer effectively to online environments without any re-training. Overall, these results confirm the robustness of UR$^2$ in noisy online environments and validate its practical deployment potential.

\subsection{Novelty and Concurrent Work}

Concurrent large-scale systems such as WebSailor-V2~\citep{li2025websailor}, Beyond Ten Turns~\citep{gao2025beyond}, and Tongyi DeepResearch~\citep{team2025tongyi} independently adopt RL-driven multi-step retrieval pipelines with \textbf{intermediate summarization}, typically relying on 32B+ models and extensive compute. Their architectural convergence suggests that coordinating retrieval and reasoning under RL is a \emph{structural challenge} rather than an ad-hoc engineering choice. Our contribution is to systematically study and address this challenge, specifically the retrieval--reasoning imbalance, under limited model capacity, showing that effective coordination can be learned even with 3--8B models. As shown in Table~\ref{tab:retrieval_freq}, UR$^2$ learns task-adaptive retrieval strategies: invoking retrieval sparingly for math (0.94 calls/query) but frequently for knowledge-intensive multi-hop QA (2.46 calls/query), with detailed analysis in Appendix~\ref{sec:retrieval_behavior}. Unlike these concurrent systems that require extensive infrastructure, UR$^2$ achieves comparable or superior performance with a single training run of ${\sim}$160 GPU-hours on 8$\times$A100 GPUs, making the approach accessible to researchers with limited compute budgets.

\section{Conclusion}

We presented UR$^2$, a unified framework that integrates retrieval-augmented generation with reasoning through reinforcement learning. Unlike existing RAG-RL approaches limited to specific domains, UR$^2$ demonstrates versatility across mathematical reasoning, medical QA, and open-domain tasks. Our key innovations---difficulty-aware curriculum learning and LLM-summarized retrieval---enable dynamic retrieval-reasoning coordination by learning \emph{when} and \emph{what} to retrieve based on problem difficulty, while preserving native reasoning capabilities. Extensive experiments on Qwen-2.5-3B/7B and LLaMA-3.1-8B show that UR$^2$ consistently outperforms advanced RAG and RL baselines, with 7B models achieving performance comparable to GPT-4o-mini and GPT-4.1-mini on several benchmarks. The framework's robustness across online and offline retrieval settings, diverse corpus configurations, and multiple summarization models confirms its practical applicability. UR$^2$ represents a significant step toward adaptive AI systems that flexibly combine parametric knowledge with dynamic information access.

\section*{Acknowledgements}

This work is supported by the National Natural Science Foundation of China (62372260, 62276152), and Wuxi Research Institute of Applied Technologies, Tsinghua University. Weizhi Ma is also supported by Beijing Nova Program.

\section*{Limitations}
While UR$^2$ shows strong performance across diverse tasks, some limitations remain. First, our primary experiments use models up to 8B parameters. A preliminary experiment on Qwen-2.5-14B-Instruct using medical training data without summarization (top-$k$=2) shows that UR$^2$ still triggers retrieval effectively, achieving 72.4\% on MedQA versus 69.8\% for CoT, suggesting the framework scales to larger models; however, comprehensive evaluation at larger scales remains for future work. Second, our reliance on on-the-fly LLM-summarized corpora may not fully reflect the complexity of raw web content. Despite these issues, UR$^2$ achieves substantial gains (up to 29.6\% improvement) and generalizes well across domains.

Future work will explore scaling UR$^2$ to 14--32B parameters with full summarization support, incorporating online corpora during training to better capture real-world retrieval dynamics, and investigating more efficient training strategies to reduce costs.

\bibliography{custom}

\appendix

\section{Use of AI Tools}
In preparing this work, we used commercial LLMs (e.g., Claude 4.0) for non-creative assistance such as language polishing, formatting, and minor code edits. These tools were not involved in method design, experimental setup, or any substantive creative contribution.

\section{Additional Experimental Results}

\subsection{Additional Ablation Results}
\label{sec:additional_ablation}

\begin{table}[h!]
\centering
\begin{minipage}[t]{0.49\textwidth}
\centering
\small
\setlength{\tabcolsep}{0pt}
\begin{tabular}{l cccc}
\toprule 
{\textbf{Method}} 
& {\textbf{Hotpot$^\dag$}} 
& {\textbf{2Wiki$^\dag$}} 
& {\textbf{Bamb.$^\ddag$}} 
& {\textbf{MusiQ.$^\ddag$}}  \\
\midrule
\textbf{UR$^2$} & \textbf{71.2} & \textbf{62.6} & \textbf{64.5} & \textbf{35.8}  \\
\midrule
w/o Math Data &70.9& 61.2& 63.3 & 34.4 \\
\quad w/o LLM Summary  & 71.0 & 62.0 & 62.7 & 33.8  \\
\midrule
weaker Stage-1 & 69.5 & 62.2  & 59.0 & 34.4  \\
\bottomrule
\end{tabular}
\caption{
Ablation Study of Qwen-2.5-7B-Instruct on open-domain QA. We report F1 scores (in \%) here. The second variant removes LLM Summary on top of the first variant without Math Data.
}
\label{tab:ablation_qa}
\end{minipage}%
\end{table}

We further conduct ablation studies on UR$^2$ in open-domain QA tasks (Table~\ref{tab:ablation_qa}).The \textit{w/o Math Data} variant shows minimal impact (0.3-1.4\% drops), confirming multi-task training preserves QA performance. Additionally removing LLM Summary causes larger drops on out-of-domain tasks (2.0\% on MusiQue) while maintaining in-domain performance, indicating LLM-summarized corpus benefits generalization. The \textit{weaker Stage-1} variant shows the largest degradation on Bamboogle (5.5\% drop), highlighting proper retrieval 

\setlength{\tabcolsep}{3pt} 
\begin{table*}[h!]
\centering
\small
\begin{tabular}{l cccc cc}
\toprule
{\textbf{Method}} & {\textbf{HotpotQA$^\dag$}} & {\textbf{2Wiki$^\dag$}} & {\textbf{Bamboogle$^\ddag$}} & {\textbf{MusiQue$^\ddag$}} & {\textbf{Math$^\dag$}} & {\textbf{Minerva$^\ddag$}} \\
\midrule
Raw Data &  72.3 &  61.9 &  58.9 &  36.1 &  79.0 &  60.5\\
Filtered Data &  71.0 &  62.0 &  62.7 &  33.8 &  78.2 &  59.4\\
\bottomrule
\end{tabular}
\caption{
Training Data Ablation for Qwen-2.5-7B-Instruct (Vanilla RL on Open-domain QA w/o Summary and Math Tasks). We report F1 scores (\%) for open-domain QA.}
\label{tab:ablation_data}
\end{table*}

\begin{table*}[h!]
\centering
\small
\setlength{\tabcolsep}{5pt} 
\begin{tabular}{l ccccc c@{\hspace{0.2em}}cc}
\toprule
\multirow{2}{*}{\textbf{Method}} 
& \multicolumn{5}{c}{\textbf{MMLU-Pro}} 
& \multicolumn{3}{c}{\textbf{Medicine}} \\
\cmidrule(lr){2-6} \cmidrule(lr){7-9} 
& Hist.$^\dag$ & Phil.$^\dag$ & Econ.$^\dag$ & Law$^\ddag$ & Avg 
& MedQA$^\dag$ & M-Med$^\ddag$ & Avg \\
\midrule
\multicolumn{9}{c}{\textbf{{\textit{Qwen-2.5-7B}}}} \\
Vanilla RL MCQ &47.2 &\underline{46.1} &61.8 &33.0 &47.0 &\underline{65.6} &\underline{60.0} &\underline{62.8} \\
Vanilla RL &\underline{52.2}& 43.5& \underline{64.0}& \underline{33.8}& \underline{48.4}& 64.2& 57.4& 60.8\\
\textbf{UR$^2$} &\textbf{53.2}& \textbf{53.0}& \textbf{72.2}& \textbf{35.0}& \textbf{53.3}& \textbf{69.6}& \textbf{62.8}& \textbf{65.9}\\
\midrule
\multicolumn{9}{c}{\textbf{{\textit{Qwen-2.5-3B}}}} \\
Vanilla RL MCQ &\underline{42.3} &\underline{37.1} &\underline{57.4} &\underline{25.0} &\underline{40.6} &50.2 &45.0 &47.6 \\
Vanilla RL &40.7 &34.7 &55.0 &24.6 &38.7 &\underline{51.8} &\underline{47.6} &\underline{49.7}\\
\textbf{UR$^2$} & \textbf{47.8} & \textbf{49.3} & \textbf{63.9} & \textbf{30.0} & \textbf{47.8} & \textbf{59.8} & \textbf{56.8} & \textbf{58.3} \\

\bottomrule
\end{tabular}
\caption{
Ablation study of Vanilla RL on Qwen-2.5-7B-Instruct and Qwen-2.5-3B-Instruct across multiple-choice reasoning tasks.
}
\label{tab:reasoning_extra_data}
\end{table*}

\noindent initialization is crucial for complex multi-hop reasoning. These results validate our design choices contribute meaningfully across diverse task types.

Table~\ref{tab:ablation_data} validates the effectiveness of our difficulty-aware data selection strategy. Despite using significantly less training data, the filtered dataset achieves comparable or even superior performance, particularly on out-of-domain tasks (e.g., Bamboogle improves from 58.9\% to 62.7\%). This indicates that reinforcement learning benefits more from high-quality, difficulty-balanced samples than from large-scale unfiltered data. As a result, UR$^2$ enables computationally efficient training while maintaining strong generalization across diverse reasoning tasks.

To ensure a fair comparison, we evaluate Vanilla RL MCQ (MMLU-Pro and Medicine tasks), which trains on mixed multiple-choice tasks similar to UR$^2$. As shown in Table~\ref{tab:reasoning_extra_data}, Vanilla RL MCQ exhibits task-dependent performance: on Qwen-2.5-7B it improves Medicine performance (62.8\% vs. 60.8\%) but lowers MMLU-Pro scores (47.0\% vs. 48.4\%), with the reverse trend on 3B models. Despite these gains, UR$^2$ consistently outperforms both Vanilla RL variants across all domains and scales, achieving average improvements on MMLU-Pro of 5.9\% for 7B models and 9.1\% for 3B models, confirming that its advantage arises from the unified retrieval-reasoning framework rather than task mixing alone.

\begin{table}
\begin{minipage}[t]{0.49\textwidth}
\centering
\small
\begin{tabular}{l cccc c}
\toprule
\multirow{2}{*}{\textbf{Summarizer}} 
& \multicolumn{4}{c}{\textbf{MMLU-Pro (EM \%)}} 
& \multirow{2}{*}{\textbf{AVG}} \\
\cmidrule(lr){2-5}
& Hist.$^\dag$ & Phil.$^\dag$ & Econ.$^\dag$ & Law$^\ddag$ \\
\midrule
GPT-4.1     & \underline{53.2} & \textbf{53.1} & \textbf{72.2} & \textbf{35.0} & \textbf{53.4} \\
Qwen-3-32B    & 52.5 & 50.9 & \underline{72.0} & 33.6 & \underline{52.3} \\
Qwen-3-8B  & 52.5   & 51.5  & 71.0   & 32.8   & 52.0 \\
GPT-4.1-mini   & 52.5 & 51.3 & 69.4 & 33.6 & 51.7\\
GPT-4o-mini & 51.8   & 49.1   & 67.4   & \underline{34.5}   & 50.8 \\
Distilled-Qwen-2.5-7B &  49.6   & \underline{52.3}   & 68.2   & 31.4  & 50.4 \\
Qwen-2.5-7B-Instruct &  \textbf{53.4}   & 47.4   & 67.2   & 32.0  & 50.0 \\
w/o Summary       & 52.1   & 48.3   & 68.0   & 32.2   & 50.2 \\
\midrule
Vanilla RL &52.2 &43.5 &64.0 &33.8 &48.4\\
\bottomrule
\end{tabular}
\caption{
Ablation on summarizers in UR$^2$ (Qwen-2.5-7B-Instruct) on MMLU-Pro. ``w/o Summary'' uses top-$3$ documents without summarizing; ``Qwen-3-32B'' uses top-$16$ documents; ``Qwen-2.5-7B'' (instruct) uses  top-$5$ documents.
}
\label{tab:summary_sources}
\end{minipage}
\end{table}

Overall, the ablations confirm that Stage-1 initialization is crucial for complex reasoning, difficulty-aware filtering yields better performance with fewer samples, and task mixing improves efficiency without accuracy loss. Importantly, \textbf{LLM-summarized retrieval highlights the necessity of addressing retrieval noise in RAG-RL methods}, guiding more stable and generalizable reasoning.

\subsection{Impact of LLM Summary and Corpus on UR$^2$ Performance}
\label{sec:corpus_ablation}

Table~\ref{tab:summary_sources} examines the robustness of UR$^2$ across different LLM summary sources. Remarkably, our framework maintains strong performance regardless of the summarization model quality. While GPT-4.1 achieves the best results (53.4\% average), even using smaller open-source models like Qwen-3-8B (52.0\%) or budget-friendly APIs like GPT-4o-mini (50.8\%) yields substantial improvements over Vanilla RL (48.4\%). Most notably, the \textit{w/o Summary} variant still achieves 50.2\%--demonstrating that our two-stage training and retrieval-aware prompting mechanisms are inherently robust and 

\begin{table*}[h!]
\small
\centering
\setlength{\tabcolsep}{1.2pt}
\begin{tabular}{c c l ccc ccc ccc ccc ccc}
\toprule 
\multirow{2}{*}{\textbf{Corpus}} & \multirow{2}{*}{\textbf{Summ.}}& \multirow{2}{*}{\textbf{Models}} & \multicolumn{3}{c}{\textbf{Hotpot$^\dag$}} & \multicolumn{3}{c}{\textbf{2Wiki$^\dag$}} & \multicolumn{3}{c}{\textbf{Bamb.$^\ddag$}} & \multicolumn{3}{c}{\textbf{MusiQ.$^\ddag$}} & \multicolumn{3}{c}{\textbf{Avg}} \\
\cmidrule(lr){4-6} \cmidrule(lr){7-9} \cmidrule(lr){10-12} \cmidrule(lr){13-15} \cmidrule(lr){16-18}
&&  & F1 & LSJ & \#R & F1 & LSJ &\#R & F1 & LSJ&\#R & F1 & LSJ &\#R & F1 & LSJ &\#R \\
\midrule
\multirow{4}{*}{\textbf{Abs}}&\multirow{4}{*}{\textbf{\ding{51}}} & ZeroSearch &46.0 &50.4 &0.66 &38.4 &38.6 &0.73 &35.8 &38.4 &0.54 &14.7 &13.8 &0.62 &33.7 &35.3 &0.64 \\
&& Search-R1 &\textbf{72.4} &78.8 &1.92 &61.0 &63.8 &3.16 &58.9 &56.8 &2.58 &32.2 &32.0 &2.92 &56.1 &57.9 &2.64 \\
&& R1-Searcher &71.8 &78.0 &1.93 &57.9 &63.6 &2.17 &56.5 &53.6 &2.02 &33.2 &32.6 &2.33 &54.9 &57.0 &2.11 \\
&& \textbf{UR$^2$} &71.2 &\textbf{79.4} &2.22 &\textbf{62.6} &\textbf{65.0} &2.72 &\textbf{64.5} &\textbf{62.4} &2.30 &\textbf{35.8} &\textbf{34.6} &2.61 &\textbf{58.5} &\textbf{60.4} &2.46 \\
\midrule
\multirow{4}{*}{\textbf{Abs}} &\multirow{4}{*}{\textbf{\ding{55}}} & ZeroSearch &44.1 &47.0 &0.64 &32.9 &31.8 &0.66 &32.6 &35.2 &0.52 &14.3 &11.8 &0.61 &31.0 &31.5 &0.61 \\
&& Search-R1 &65.8 &72.4 &2.68 &41.8 &51.6 &3.54 &44.8 &44.8 &2.96 &25.1 &24.1 &3.49 &44.4 &48.2 &3.17 \\
&& R1-Searcher &\textbf{69.7} &\textbf{75.2} &2.16 &56.6 &58 &2.45 &41.7 &40.0 &2.38 &23.7 &22.4 &2.84 &47.9 &48.9 &2.46 \\
&& \textbf{UR$^2$} &67.6 &73.6 &1.98 &\textbf{59.1} &\textbf{59.6} &2.53 &\textbf{47.5} &\textbf{47.2} &2.10 &\textbf{28.2} &\textbf{25.4} &2.43 &\textbf{50.6} &\textbf{51.5} &2.26 \\
\midrule
\multirow{4}{*}{\textbf{Full}}&\multirow{4}{*}{\textbf{\ding{51}}} & ZeroSearch &44.3 &48.8 &0.74 &36.5 &36.8 &0.90 &46.3 &44.8 &0.70 &19.3 &20.0 &0.81 &36.6 &37.6 &0.79 \\
&& Search-R1 &\textbf{66.0} &67.2 &2.01 &60.6 &65.6 &3.12 &70.0 &71.2 &2.06 &37.8 &39.0 &2.69 &58.6 &60.8 &2.47 \\
&& R1-Searcher &62.9 &\textbf{68.0} &1.97 &62.5 &\textbf{66.8} &2.15 &69.0 &65.6 &1.86 &36.7 &37.8 &2.24 &57.8 &59.6 &2.06 \\
&& \textbf{UR$^2$} &62.6 &\textbf{68.0} &2.11 &\textbf{63.3} &67.6 &2.73 &\textbf{73.0} &\textbf{74.0} &2.13 &\textbf{40.4} &\textbf{42.0} &2.55 &\textbf{59.8} &\textbf{62.9} &2.38 \\
\midrule
\multirow{4}{*}{\textbf{Full}}&\multirow{4}{*}{\textbf{\ding{55}}}& ZeroSearch &39.2 &41.6 &0.63 &34.0 &33.8 &0.67 &34.1 &36.0 &0.50 &13.4 &11.8 &0.58 &30.2 &30.8 &0.59 \\
&& Search-R1 &\textbf{57.4} &60.6 &2.75 &49.2 &51.0 &3.50 &57.6 &55.2 &2.82 &26.9 &26.4 &3.40 &47.8 &48.3 &3.12 \\
&& R1-Searcher &57.6 &\textbf{61.6} &2.24 &\textbf{56.0} &\textbf{59.0} &2.37 &\textbf{57.5} &\textbf{57.6} &2.07 &26.8 &\textbf{26.6} &2.63 &\textbf{49.5} &\textbf{51.2} &2.33 \\
&& \textbf{UR$^2$} &54.6 &60.6 &2.03 &54.5 &55.8 &2.51 &52.6 &49.6 &2.06 &\textbf{27.8} &26.2 &2.38 &47.4 &48.1 &2.25 \\
\bottomrule
\end{tabular}
\caption{Performance of UR$^2$ and baselines on open-domain QA datasets across different corpus configurations. 
\texttt{Abs} denotes corpora based on Wikipedia abstracts, while \texttt{Full} uses full articles. For each corpus, we use top-$10$ documents with summaries and top-$5$ without. 
\#R represents the number of successful retrievals per question.
}
\label{tab:open_domain_qa_1}
\end{table*}

\noindent not dependent on expensive summarization models. This flexibility makes UR$^2$ practically deployable across various computational budgets while maintaining its effectiveness, confirming the generalizability of our approach beyond specific model configurations. 

Table~\ref{tab:open_domain_qa_1} investigates the impact of different corpus configurations on UR$^2$'s performance across open-domain QA tasks. The results reveal several key insights about corpus design choices. First, using Wikipedia abstracts (\texttt{Abs}, released with HotpotQA) versus full articles (\texttt{Full}) shows task-dependent effects: abstracts perform better on easy questions (HotpotQA), while full articles excel on complex reasoning tasks requiring broader context (2Wiki, Bamboogle, MusiQue). Second, the presence of LLM summarization consistently improves performance across all configurations, with average F1 scores increasing by 6.5-10.8\% when summaries are applied. Notably, UR$^2$ maintains competitive performance even without summaries (50.6\% F1 with \texttt{Abs}, 47.4\% with \texttt{Full}), substantially outperforming ZeroSearch's reliance on synthetic content. The retrieval frequency (\#R) analysis shows that UR$^2$ strategically balances retrieval calls---using fewer retrievals than Search-R1 while achieving superior performance, demonstrating more efficient knowledge utilization.

Table~\ref{tab:summary_med} examines corpus selection for domain-specific tasks, comparing general Wikipedia against specialized MedQA textbooks for medical reasoning. The results demonstrate that domain-specific corpora provide marginal improvements when summarization is applied (70.2\% vs. 69.6\% on MedQA), but this advantage diminishes without summaries. More importantly, the performance gap between summarized and non-summarized variants is substantial (8.4\% on MedQA with Wikipedia), highlighting that effective summarization is more critical than corpus specialization. This finding suggests that UR$^2$'s LLM-summarized approach can effectively bridge the gap between general and specialized knowledge sources, making it practical for deployment across diverse domains without extensive corpus curation.

\setlength{\tabcolsep}{5pt}
\begin{table}[h!]
\centering
\begin{tabular}{l c cc}
\toprule
\textbf{Corpus} & \textbf{Summ.} & \textbf{Medicine$^\dag$} & \textbf{M-Med$^\ddag$} \\
\midrule
Wikipedia & \ding{51} & 69.6 & 62.8 \\
Textbooks & \ding{51} & 70.2 & 63.8 \\
\midrule
Wikipedia & \ding{55} &61.2 &59.2 \\
Textbooks & \ding{55} &62.0 &58.0 \\
\bottomrule
\end{tabular}
\caption{
Ablation study of UR$^2$ on the medical reasoning tasks. We compare different corpus (Wikipedia vs.~MedQA Textbooks) and the effect of applying summarization. ``w/o Summary'' uses top-3 retrieved document.
}
\label{tab:summary_med}
\end{table}

Collectively, Tables~\ref{tab:summary_sources}, ~\ref{tab:open_domain_qa_1}, and~\ref{tab:summary_med} demonstrate robustness of UR$^2$ across three dimensions: corpus configuration, domain specialization, and summarization quality. The framework maintains strong performance whether using abstracts or full articles, general or specialized corpora, and expensive o budget-friendly summarizers. Most remarkably,

\begin{figure*}[h]
    \centering
    \includegraphics[width=2\columnwidth]{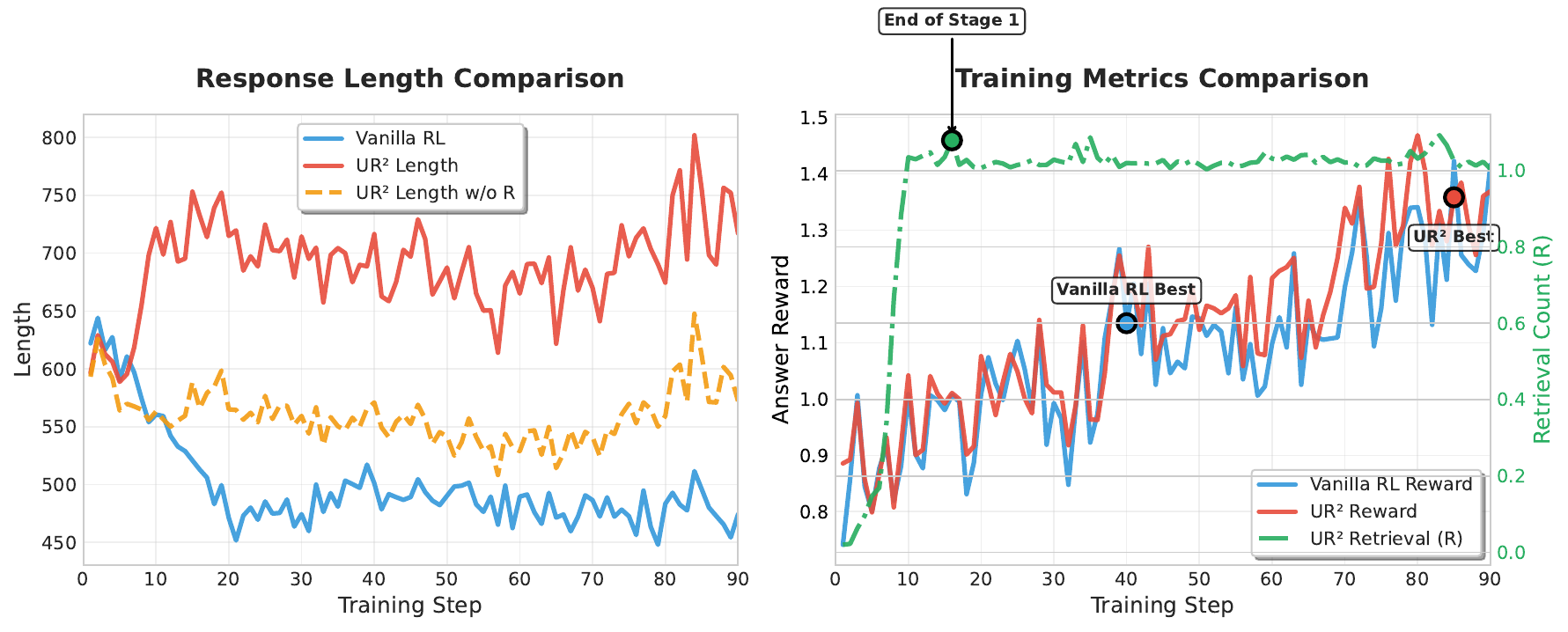}
    \caption{Comparison of Vanilla RL MCQ and UR$^2$ performance on Qwen-2.5-3B-Instruct across training steps. Peak test set performances are indicated.}
    \label{fig:rl_ur2_comparison}
\end{figure*}

 \noindent even without any summarization, UR$^2$ achieves competitive results through its two-stage training and difficulty-aware retrieval mechanisms. This comprehensive ablation validates that UR$^2$'s effectiveness stems from its fundamental architecture rather than dependency on specific external resources, confirming its practical applicability across diverse computational and domain constraints.

\subsection{Comparative Analysis of Retrieval Integration in RL Training}
\label{sec:training_analysis}
Figure~\ref{fig:rl_ur2_comparison} reveals key differences between UR$^2$ and Vanilla RL MCQ on Qwen-2.5-3B-Instruct in training. Vanilla RL saturates early at step 40 with 1.1 reward, with later gains mainly due to repeated data every 47 steps. In contrast, UR² steadily improves to 1.4 reward by step 85, matching the 15.7\% relative benchmark gain. Retrieval frequency remains dynamic after Stage 1, showing selective use. UR² also generates longer outputs post-training, indicating deeper reasoning. This extended training capability demonstrates that retrieval-augmented approaches fundamentally expand model capacity limits, enabling continuous learning beyond traditional RL saturation points.

\subsection{Analysis of Retrieval Behavior}
\label{sec:retrieval_behavior}

We examine how UR$^2$ adapts its retrieval frequency across tasks of varying nature and difficulty. Table~\ref{tab:retrieval_freq} reports the average number of retrieval calls per query for UR$^2$ Qwen-2.5-7B across different task families.

\begin{table}[h]
\centering
\small
\begin{tabular}{lc}
\toprule
\textbf{Task} & \textbf{Avg. \#Retrieval / Query} \\
\midrule
Math500 & 0.94 \\
MMLU-Pro & 1.34 \\
Open-domain QA & 2.46 \\
\bottomrule
\end{tabular}
\caption{Average retrieval frequency per query for UR$^2$ Qwen-2.5-7B across task families. UR$^2$ learns to invoke retrieval adaptively based on task characteristics.}
\label{tab:retrieval_freq}
\end{table}

The retrieval frequency pattern is consistent with task characteristics: mathematical reasoning relies more on internal computation, while multi-hop QA requires frequent external evidence. Within open-domain QA, retrieval frequency also varies by dataset difficulty: on 2WikiMultiHopQA, which is constructed directly from Wikipedia, retrieval usage remains consistently high across all corpus settings (Table~\ref{tab:open_domain_qa_1}), whereas simpler benchmarks trigger fewer calls. Figure~\ref{fig:rl_ur2_comparison} further shows that retrieval frequency evolves dynamically during training, stabilizing after Stage~1 as the model learns to selectively invoke retrieval. These results demonstrate that UR$^2$ acquires adaptive retrieval strategies rather than relying on fixed heuristics.

\subsection{Unsuccessful Attempts on Reasoning Models}
\label{sec:unsuccessful_attempts}
We also conducted experiments on the R1-like model \texttt{DeepSeek-R1-Distill-Qwen-7B\footnote{https://huggingface.co/deepseek-ai/DeepSeek-R1-Distill-Qwen-7B}}. However, when applying the MMLU-Pro prompting setup, we observed that the model lacked any retrieval capability. This remained true even after replacing the original searching special tags with alternative tokens \texttt{<search></search>} and \texttt{<information></information>}, which were shown in the ablation study (Table~\ref{tab:mcqablation}) to more effectively trigger retrieval. These results indicate a degradation of tool-usage ability after extensive chain-of-thought training. Due to computational constraints, we did not extend training to more updated models such as the Qwen-3 series. We plan to supplement this work with relevant code and experiments in future updates.

\subsection{Compression Ratio of the Summarization Module}
\label{sec:compression_ratio}

To quantitatively evaluate the effectiveness of the summarization module, we analyze its compression behavior under both offline and online retrieval settings. Specifically, we randomly sampled 10,000 entries from offline retrieval intermediate files and computed token-level statistics before and after summarization using the Qwen-2.5-7B-Instruct tokenizer.

For offline retrieval, the average number of tokens is reduced from 1,291 to 128 after summarization, corresponding to a compression ratio of \textbf{7.87$\times$}. Although offline retrieval already operates on relatively cleaner and shorter documents, the summarization module still removes a substantial amount of redundant or weakly relevant information, yielding a significantly more compact context for downstream reasoning.

We further conduct the same analysis in a more challenging online retrieval setting, where retrieved documents are typically longer and noisier. Using 3,000 randomly sampled online retrieval instances, we observe that the average token length is reduced from 11,346 to 761, resulting in a much higher compression ratio of \textbf{14.91$\times$}. This stark contrast highlights the growing importance of the summarization module as retrieval environments become more realistic and complex.

Overall, these results demonstrate that the summarization module consistently achieves substantial compression across different retrieval regimes, while its impact is particularly pronounced in online retrieval scenarios. This capability is crucial for controlling context length, reducing computational overhead, and enabling efficient multi-step search--reasoning under real-world conditions.

\subsection{Difficulty-Aware Curriculum Preserves Intrinsic Reasoning Ability}
\label{sec:reasoning_preservation}
We explicitly design the difficulty-aware curriculum to preserve, rather than weaken, the model's intrinsic reasoning ability. During training, for samples on which the model achieves an accuracy of at least 0.5, we select direct reasoning instead of retrieval-augmented generation. This strategy serves two purposes: (1) it avoids unnecessary retrieval calls and reduces API costs, and (2) it ensures that the model continues to practice and retain pure reasoning skills in domains where retrieval is not essential.

As shown in Table~\ref{tab:reasoning_preservation} (Section~\ref{sec:ablation}), UR$^2$ consistently outperforms Vanilla-RL on most benchmarks, even when evaluated using direct reasoning prompts. These results demonstrate that difficulty-aware curriculum learning does not degrade intrinsic reasoning ability. Instead, it strengthens the synergy between retrieval usage and standalone reasoning performance.

\subsection{Comprehensive Supplementary Results on Open-Domain and Reasoning Tasks}
\label{sec:comprehensive_baselines}

Tables~\ref{tab:reasoning_extra} and~\ref{tab:open_domain_qa_2} provide supplementary experimental results, focusing on Advanced RAG methods across different model scales and GPT-4o-mini performance on open-domain QA tasks.

\begin{table*}[ht!]
\centering
\small
\setlength{\tabcolsep}{1.5pt} 
\begin{tabular}{l ccccc c@{\hspace{0.2em}}cc c@{\hspace{0.2em}}cc}
\toprule
\multirow{2}{*}{\textbf{Method}} 
& \multicolumn{5}{c}{\textbf{MMLU-Pro}} 
& \multicolumn{3}{c}{\textbf{Medicine}} 
& \multicolumn{3}{c}{\textbf{Math}} \\
\cmidrule(lr){2-6} \cmidrule(lr){7-9} \cmidrule(lr){10-12}
& Hist.$^\dag$ & Phil.$^\dag$ & Econ.$^\dag$ & Law$^\ddag$ & Avg 
& MedQA$^\dag$ & M-Med$^\ddag$ & Avg 
& Math500$^\dag$ & Minerva$^\ddag$ & Avg \\
\midrule
\multicolumn{12}{c}{\textbf{\textit{Qwen-2.5-7B}}} \\
\multicolumn{12}{c}{\fontsize{7.5pt}{9pt}\selectfont\textcolor{gray}{\textit{Advanced RAG Methods}}} \\
Self-Ask &40.7 &42.1 &60.0 &26.2 &42.3 &51.8 &47.8 &49.8 &74.9 &57.7 &66.3  \\
RAT &47.2 &44.7 &64.4 &30.0 &46.6 &60.0 &53.2 &56.6 &74.4 &55.5
&65.0  \\
Search-o1 &42.8 &45.9 &63.2 &29.6 &45.4 &58.2 &52.8 &55.6 &78.2 &60.3 &69.3  \\

\multicolumn{12}{c}{\textbf{{\textit{Qwen-2.5-3B}}}} \\
CoT &33.6 &32.3 &48.8 &20.6 &33.8 &39.4 &36.8 &38.1 &63.6 &39.9 &51.8  \\
Standard RAG &37.8 &\underline{36.5} &51.4 &23.2 &37.1 &45.6 &40.0 &42.8 &65.3 &40.8 &53.1  \\
\multicolumn{12}{c}{\fontsize{7.5pt}{9pt}\selectfont\textcolor{gray}{\textit{Advanced RAG Methods}}} \\
Self-Ask &33.1 &30.5 &44.2 &20.2 &32.0 &39.6 &36.0 &37.8 &58.3 &40.3 &49.3  \\
RAT &37.8 &33.5 &50.6 &18.4 &35.1 &47.4 &42.6 &45.0 &67.9 &\underline{44.9} &\underline{56.4}  \\
Search-o1 &33.9 &34.5 &50.6 &20.6 &34.9 &44.6 &37.4 &41.0 &\textbf{69.4} &43.0 &56.2  \\
\multicolumn{12}{c}{\fontsize{8.0pt}{9.5pt}\selectfont\textcolor{gray}{\textit{Our Implementations}}} \\
Vanilla RL &\underline{40.7} &34.7 &\underline{55.0} &\underline{24.6} &\underline{38.7} &\underline{51.8} &\underline{47.6} &\underline{49.7} &68.0 &43.9 &52.4  \\
\textbf{UR$^2$} & \textbf{47.8} & \textbf{49.3} & \textbf{63.9} & \textbf{30.0} & \textbf{47.8} & \textbf{59.8} & \textbf{56.8} & \textbf{58.3} & \textbf{69.4} & \textbf{45.0} & \textbf{57.2}   \\
\midrule

\multicolumn{12}{c}{\textbf{{\textit{LLaMA-3.1-8B}}}} \\
CoT &37.8 &\textbf{40.9} &\underline{53.4} &\textbf{29.0} &\underline{40.3} &59.6 &52.6 &56.1 &48.4 &34.4 &41.4  \\
Standard RAG &\underline{43.6} &33.9 &51.0 &26.6 &38.8 &56.4 &53.2 &54.8 &45.0 &31.4 &38.2  \\
\multicolumn{12}{c}{\fontsize{7.5pt}{9pt}\selectfont\textcolor{gray}{\textit{Advanced RAG Methods}}} \\
Self-Ask &39.8 &32.1 &47.0 &23.4 &35.6 &53.0 &42.8 &47.9 &46.9 &27.0 &37.0  \\
RAT &42.3 &37.7 &52.6 &28.6 &40.3 &63.8 &56.0 &59.9 &\underline{50.1} &36.8 &43.5  \\
Search-o1 &32.6 &32.5 &46.0 &28.0 &34.8 &56.0 &46.0 &56.6 &41.5 &27.8 &34.7  \\
\multicolumn{12}{c}{\fontsize{8.0pt}{9.5pt}\selectfont\textcolor{gray}{\textit{Our Implementations}}} \\
Vanilla RL &44.6 &36.9 &53.0 &26.4 & 40.2 &\underline{66.8} &\underline{57.4} &\underline{62.1} &45.5 &\textbf{43.4} &\underline{44.4}  \\
\textbf{UR$^2$} &\textbf{48.3} &\underline{38.6} &\textbf{58.0} &\underline{28.8} & \textbf{43.4} &\textbf{68.6} &\textbf{58.4} &\textbf{63.5} &\textbf{54.5} &\underline{39.0} &\textbf{46.8} \\

\bottomrule
\end{tabular}
\caption{Extended results on GPT-4o-mini, Qwen-2.5-3B-Instruct, and LLaMA-3.1-8B-Instruct across reasoning tasks.  We report EM scores (\%) for MMLU-Pro and MedQA, and \texttt{LLM-as-a-judge} scores (\%) for math benchmarks. $\dag$ = in-domain; $\ddag$ = out-of-domain.}
\label{tab:reasoning_extra}
\end{table*}

The extended results reveal significant performance limitations of Advanced RAG methods for open-source models. On Qwen-2.5-3B, Self-Ask achieves only 32.0\% on MMLU-Pro, substantially underperforming even basic CoT (33.8\%). RAT shows inconsistent performance, achieving competitive results on medical tasks (45.0\%) but poor performance on Law (18.4\%), indicating fragility in cross-domain generalization. Search-o1 demonstrates moderate effectiveness, reaching 41.0\% on medical tasks, but fails to achieve consistent improvements across reasoning domains. On LLaMA-3.1-8B, Advanced RAG methods exhibit mixed results. While RAT achieves reasonable performance on Medicine (59.9\%) and Math (43.5\%), Self-Ask and Search-o1 show notable degradation compared to basic CoT on several sub-domains. These results highlight the challenge of scaling sophisticated retrieval mechanisms to diverse model architectures and reasoning tasks.

GPT-4o-mini establishes strong performance on open-domain QA, with Search-o1 achieving 48.9\% F1 average, significantly outperforming other Advanced RAG methods (41.3 and 41.8\%). Additionally, RAT and Self-Ask incur prohibitive API costs due to their sentence-level analysis and rewriting operations, making them impractical for large-scale deployment. Notably, Standard RAG achieves competitive performance (42.1\% F1) on GPT-4o-mini, suggesting that larger commercial models can effectively leverage simple retrieval without sophisticated coordination mechanisms. The performance gap between GPT-4o-mini (48.9\%) and smaller models, such as Qwen-2.5-3B (27.8\%) for Search-o1, highlights the substantial challenge of achieving effective retrieval-reasoning integration in resource-constrained settings and validates the necessity of our specialized framework design.

\setlength{\tabcolsep}{1.7pt}
\begin{table*}[h!]
\centering
\small
\begin{tabular}{p{1.9cm}>{\centering\arraybackslash}p{2.3cm} lcccccccccc}
\toprule 
\multirow{2}{*}{\textbf{Models}} & \multirow{2}{*}{\textbf{Types}} & \multirow{2}{*}{\textbf{Methods}} & \multicolumn{2}{c}{\textbf{Hotpot$^\dag$}} & \multicolumn{2}{c}{\textbf{2Wiki$^\dag$}} & \multicolumn{2}{c}{\textbf{Bamb.$^\ddag$}} & \multicolumn{2}{c}{\textbf{MusiQ.$^\ddag$}} & \multicolumn{2}{c}{\textbf{Avg}} \\
\cmidrule(lr){4-5} \cmidrule(lr){6-7} \cmidrule(lr){8-9} \cmidrule(lr){10-11} \cmidrule(lr){12-13}
& & & F1 & LSJ & F1 & LSJ & F1 & LSJ & F1 & LSJ & F1 & LSJ \\
\midrule
\multirow{5}{*}{\textbf{GPT-4o-mini}} & \multirow{2}{*}{Vanilla Methods} & CoT &46.5 &51.2 &35.0 &35.4 &\underline{55.2} &\textbf{62.4} &24.9 &26.8 &40.4 &44.0 \\
&& Standard RAG & \underline{59.6} & \underline{69.6} & \underline{43.0} & \underline{45.8} & 46.7 & 46.4 & 19.3 & 21.6 & \underline{42.1} & \underline{45.9} \\
\cmidrule{3-3}
&\multirow{3}{*}{Advanced RAG} & Self-Ask &45.0 &50.4 &36.9 &40.0 &\textbf{59.3} &\underline{57.6} &\underline{26.1} &\underline{27.8} &41.8 &44.0 \\
&& RAT &53.8 &59.2 &34.1 &34.8 &53.0 &51.2 &24.3 &24.8 &41.3 &42.5 \\
&& Search-o1 &\textbf{64.3} &\textbf{73.4} &\textbf{47.3} &\textbf{52.0} &54.5 &56.0 &\textbf{29.6} &\textbf{30.2} &\textbf{48.9} &\textbf{52.9} \\
\midrule
\multirow{11}{*}{\textbf{Qwen-2.5-7B}} & \multirow{2}{*}{Vanilla Methods} & CoT &24.9 &31.0 &25.1 &27.6 &41.3 &43.2 &14.8 &12.2 &26.5 &28.5 \\
&& Standard RAG &49.2 &62.8 &32.8 &37.6 &38.9 &40.0 &14.4 &14.6 &33.8 &38.8 \\
\cmidrule{3-3}
&\multirow{3}{*}{Advanced RAG} & Self-Ask &28.8 &61.0 &22.2 &45.4 &28.9 &42.4 &13.6 &19.6 &23.4 &42.1 \\
&& RAT &37.9 &40.6 &23.3 &23.6 &31.6 &30.4 &14.4 &12.4 &26.8 &26.8 \\
&& Search-o1 &50.9 &61.6 &45.2 &48.6 &37.5 &39.2 &20.6 &19.8 &38.6 &42.3 \\
\cmidrule{3-3}
&\multirow{4}{*}{RAG-RL} & R1-Searcher &\underline{71.8} &78.0 &57.9 &63.6 &56.5 &53.6 &33.3 &32.6 &54.8 &57.0\\
&& Search-R1 &\textbf{72.4} &\underline{78.8} &61.0 &63.8 &58.9 &56.8 &32.2 &32.0 &56.1 &57.9\\
&& R1-Searcher++ &59.0 &64.2 &\underline{61.2} &\underline{64.4} &60.8 &59.2 &33.8 &32.8 &53.7 &55.2\\
&& ZeroSearch &46.0 &50.4 &38.4 &38.6 &35.8 &38.4 &14.7 &13.8 &33.7 &35.3 \\
\cmidrule{3-3}
&\multirow{2}{*}{\begin{minipage}[c]{2.5cm}\centering Our Implementations\end{minipage}} & Vanilla RL &70.9 &\underline{78.8} &\underline{61.2} &62.4 &\underline{63.3} &\textbf{63.2} &\underline{34.4} &\underline{34.4} &\underline{57.5} &\underline{59.6} \\
 && \textbf{UR$^2$} &71.2 &\textbf{79.4} &\textbf{62.6} &\textbf{65.2} &\textbf{64.5} &\underline{62.4} &\textbf{35.8} &\textbf{34.6} &\textbf{58.5} &\textbf{60.4} \\
\midrule
\multirow{9}{*}{\textbf{Qwen-2.5-3B}} & \multirow{2}{*}{Vanilla Methods} & CoT &26.6& 27.2&22.7 &22.6 &31.2 &33.6 &11.3 &9.6 &23.0 &23.3 \\
&& Standard RAG &50.6 &57.0 &29.8 &30.4 &26.1 &27.2 &9.7 &7.4 &29.1 &30.5\\
\cmidrule{3-3}
&\multirow{3}{*}{Advanced RAG} & Self-Ask &33.8 &47.2 &21.0 &28.8 &30.6 &32.0 &14.5 &14.8 &25.0 &30.7 \\
&& RAT &30.1 &32.2 &15.1 &15.4 &30.6 &28.0 &11.0 &8.2 &21.7 &21.0 \\
&& Search-o1 &36.4 &37.6 &30.8 &31.8 &31.4 &32.0 &12.5 &10.0 &27.8 &27.9 \\
\cmidrule{3-3}
&\multirow{2}{*}{RAG-RL} & Search-R1 &63.1 &69.2 &49.5 &53.4 &48.3 &48.0 &27.6 &27.8 &49.6 &47.1\\
&& Zero-Search &42.7 &45.8 &26.1 &27.6 &32.4 &31.2 &16.9 &17.0 &29.5 &30.4\\
\cmidrule{3-3}
&\multirow{2}{*}{\begin{minipage}[c]{2.5cm}\centering Our Implementations\end{minipage}} & Vanilla RL&\underline{65.9} &\underline{73.6} &\underline{54.9} &\underline{58.0} &\underline{59} &\underline{57.6} &\underline{30.0} &\underline{29.6} &\underline{52.5} &\underline{54.7} \\
&& \textbf{UR$^2$} &\textbf{67.7} &\textbf{76.0} &\textbf{55.2} &\textbf{58.6} &\textbf{57.8} &\textbf{58.4} &\textbf{30.5} &\textbf{31.6} &\textbf{55.3} &\textbf{56.2}

\\
\midrule
\multirow{8}{*}{\textbf{LLaMA-3.1-8B}} & \multirow{2}{*}{Vanilla Methods} & CoT &28.6 &31.6 &16.4&17.8 &43.0 &42.4 &9.8 &10.8 &24.5 &25.7 \\
&& Standard RAG &47.5 &54.4 &26.2 &26.4 &26.5 &28.0 &10.1 &10.2 &27.6 &29.8\\
\cmidrule{3-3}
&\multirow{3}{*}{Advanced RAG} & Self-Ask &43.0 &50.8 &27.3 &29.8 &41.5 &44.8 &16.8 &16.4 &32.2 &35.5 \\
&& RAT &44.5 &48.8 &16.4 &15.6 &39.7 &39.2 &17.0 &16.0 &29.4 &29.9 \\
&& Search-o1 &53.0 &59.4 &37.5 &38.4 &30.0 &30.4 &15.9 &16.2 &34.1 &36.1 \\
\cmidrule{3-3}
&\multirow{1}{*}{RAG-RL} & R1-Searcher &\textbf{70.8} &76.8 &59.6 &62.2 &\textbf{64.7} &62.4 &31.1 &29.4 &\textbf{56.6} &57.7\\
\cmidrule{3-3}
&\multirow{2}{*}{\begin{minipage}[c]{2.5cm}\centering Our Implementations\end{minipage}} & Vanilla RL &70.0 &\underline{77.6} &\textbf{61.2} &\textbf{64.2} &60.6 &\textbf{63.2} &\underline{32.7} &\underline{31.8} &56.1 &\underline{59.2} \\
& & \textbf{UR$^2$}&\underline{70.1} &\textbf{78.8} &\underline{60.1} &\underline{63.2} &\underline{60.7}&\textbf{63.2} &\textbf{34.3} &\textbf{34.0} &\underline{56.3} &\textbf{59.8}\\
\bottomrule
\end{tabular}
\caption{
Extended results of GPT-4o-mini, Qwen-2.5-3B, and LLaMA-3.1-8B on open-domain QA. 
We report F1 and \texttt{LLM-as-a-judge} (LSJ) scores, both in \%. 
$\dag$ denotes in-domain datasets; $\ddag$ indicates out-of-domain.
}
\label{tab:open_domain_qa_2}
\end{table*}

\section{Training Details}
\subsection{Distilled Qwen-2.5-7B Summarizer Training Setup}
\label{sec:distilled_summarizer}

To remove reliance on a strong proprietary summarization model, we distill a \texttt{Qwen-2.5-7B-Instruct} summarizer and use it consistently during both training and evaluation.

\textbf{Training Data.}
The summarizer is trained on approximately 360k intermediate summaries, including 180k samples over 2 epochs from reasoning tasks (20k per MMLU-Pro subject, 30k from Medicine QA, 30k for Math and 60k for Open-domain QA). All summaries are generated by a mixture of strong teacher models, including \texttt{GPT-4.1-mini}, \texttt{GPT-4.1}, and \texttt{Qwen-3-32B}, to ensure diversity and robustness across domains.

\textbf{Training Configuration.}
We perform supervised fine-tuning using a fixed learning rate of $1\times10^{-5}$. The training runs for 312 GPU hours on H20 GPUs (8 $\times$ 39 hours). No reinforcement learning is applied during summarizer training.

\textbf{Usage in UR$^2$.}
Unless otherwise specified, this distilled summarizer is used only for ablation experiments on reasoning tasks, where it replaces proprietary summarizers during both training and evaluation. The corresponding results are reported in the main experiments on reasoning tasks.

\subsection{Training Setting Details}
\label{sec:training_set_details}

We train UR$^2$ using the REINFORCE++ algorithm~\citep{hu2025reinforce}, a simplified variant of Proximal Policy Optimization (PPO) designed to encourage exploration. In particular, we discard the critic and omit the KL-divergence term, following previous findings~\citep{zhang2025critique,song2025r1,chen2025empirical} that excessive regularization can impede effective strategy learning in sparse-reward scenarios. We retain the PPO-style clipped surrogate objective with $\epsilon=0.2$ to stabilize policy updates.

The training objective is defined as:
\begin{equation}
\label{eq1}
J_{\text{UR}^2}(\theta) = \mathbb{E}_{x, \{y^i\}} \left[ \frac{1}{G} \sum_{i=1}^{G} \frac{1}{|y^i|} \sum_{t=1}^{|y^i|} y^i_t \cdot r_{i,t} \cdot \hat{A}_{i,t} \right]
\end{equation}
where the importance weight is:
\begin{equation}
r_{i,t} = \frac{\pi_\theta(y^i_t \mid x, y^i_{<t}; o_i)}{\pi_{\text{old}}(y^i_t \mid x, y^i_{<t}; o_i)}
\end{equation}
and the normalized advantage is:
\begin{equation}
\label{eq2}
\hat{A}_{i,t} = \textit{Norm}_{\textit{batch}} \left( \textit{Norm}_{\textit{group}} \left( R_i - b \right) \right)
\end{equation}

The advantage $\hat{A}_{i,t}$ is computed by subtracting the group-level reward baseline and applying normalization across the group and batch to improve learning stability. Here, $x$ denotes the input prompt, $\{y^i\}$ are the sampled trajectories, $o_i$ is the retrieved context and $b$ is the group-level baseline (mean of $R_i$).

To reduce overfitting to retrieved content, we adopt a retrieval masking strategy~\citep{sun2025zerosearch,song2025r1,jin2025search}, which treats retrieved external knowledge as part of the observation space rather than trainable input. This encourages the model to reason based on retrieved information without directly optimizing on it. Our implementation builds upon the REINFORCE++ baseline provided by OpenRLHF~\citep{hu2024openrlhf}.

Each prompt is rolled out $G=16$ times. We use the mean reward of each rollout group as the baseline for computing the advantage of each sample. To stabilize training, we apply a two-stage normalization scheme: normalization is first performed within each rollout group, followed by global normalization across the full batch.

Training is conducted with DeepSpeed ZeRO-2~\citep{rajbhandari2020zero} for memory efficiency. We use \texttt{gpt-4.1-mini-2025-04-14} as the summarization model during training. Token limits per generation turn are set to \texttt{3072} for math tasks, \texttt{1536} for multiple-choice questions (MCQ), and \texttt{512} for open-domain QA. Sampling parameters are fixed as \texttt{temperature = 1.0} and \texttt{top\_p = 0.9}.

We train for up to 2 epochs. In practice, most models achieve optimal performance within 1.5 epochs. Therefore, we report results from the checkpoint with the best test set performance within the first 1.5 epochs. We save checkpoints every 5 steps for single-task training and 3B models, and every 10 steps for larger-scale experiments. The specific training steps for each reported model are detailed in Table~\ref{tab:training_checkpoints} below. \textit{W/o Stage-1} variant in Table~\ref{tab:mcqablation} replaces the special tags with \texttt{<search></search>} and \texttt{<information></information>}, removing the initial retrieval capability activation stage. The \textit{Weaker Stage-1} variant in Table~\ref{tab:ablation_qa} employs a modified training protocol based on UR$^2$ Qwen-2.5-7B-Instruct for MCQ tasks, where retrieval-related rewards are only provided during the initial 10 training steps. The \textit{Qw3-8B} variant in Table~\ref{tab:mcqablation} uses Qwen-3-8B for summarization with \texttt{max\_tokens = 2048}, \texttt{temperature = 0.3}, and \texttt{top\_p = 0.7}. Specifically, the retrieval reward assigns 0.5 for single retrieval attempts and 1.0 for multiple retrievals ($\ge$2), reflecting a more conservative retrieval activation strategy than that of our proposed method.

Table~\ref{tab:training_checkpoints} summarizes the training configurations and checkpoint details across all model scales. Two key observations can be drawn:

For the task mixing strategy of multiple-choice reasoning tasks, we combine MedQA and synthetic MMLU-style data, assigning most hard questions to retrieval-augmented training while maintaining an overall 1:1 ratio between retrieval-based and direct reasoning instances.

First, for Qwen models, performance consistently improves as more training compute is introduced via our UR$^2$ method. The method's design---encouraging structured retrieval behavior---ensures that increased steps and effective epochs lead to meaningful gains across tasks.

Second, while UR$^2$ also improves performance on LLaMA-3.1-8B, training on this model is observed to be less stable. Performance tends to saturate early (e.g., low effective epochs despite higher step counts), for both Vanilla RL and UR$^2$ variants. This indicates that LLaMA-3.1-8B may require different training strategies to maintain learning dynamics over time. Future work will explore alternative foundation models and optimization schedules to improve convergence and stability.

\begin{table*}[ht!]
\centering
\small
\setlength{\tabcolsep}{1.5pt}
\renewcommand{\arraystretch}{1.3}
\begin{tabular}{l c c c c}
\toprule
\textbf{Model} & \textbf{Training Dataset} & \textbf{Dataset Size} & \textbf{Checkpoint Step} & \textbf{Training Epochs} \\
\midrule
\multicolumn{5}{c}{\textcolor{gray}{\textbf{Qwen-2.5-3B - Main Experiments}}} \\
UR$^2$-Math\&QA & Math\&QA & 6000 & 47 & 1.0 \\
UR$^2$-MCQ & MMLU\&Medqa & 9000 & 85 & 1.2 \\
Vanilla RL-Math & Math & 3000 & 15 & 0.64 \\
Vanilla RL-QA & QA & 3000 & 40 & 1.7 \\
Vanilla RL-MMLU & MMLU & 6000 & 47 & 1.0 \\
Vanilla RL-MedQA & Medqa & 3000 & 40 & 1.7 \\
Vanilla RL-MCQ & MMLU\&Medqa & 9000 & 40 & 0.57 \\
\midrule
\multicolumn{5}{c}{\textcolor{gray}{\textbf{LLaMA-3.1-8B Models - Main Experiments}}} \\
UR$^2$-Math\&QA & Math\&QA & 6000 & 30  & 0.32 \\
UR$^2$-MCQ & MMLU\&Medqa & 9000 & 30 & 0.21 \\
Vanilla RL-Math & Math & 3000 & 30 & 0.64 \\
Vanilla RL-QA & QA & 3000 & 47 & 1.0 \\
Vanilla RL-MMLU & MMLU & 6000 & 60 & 0.64 \\
Vanilla RL-MedQA & Medqa & 3000  & 30 & 0.64 \\
\midrule
\multicolumn{5}{c}{\textcolor{gray}{\textbf{Qwen-2.5-7B - Main Experiments}}} \\
UR$^2$-Math\&QA & Math\&QA & 6000 & 40 & 0.43 \\
UR$^2$-MCQ & MMLU\&Medqa & 9000 & 100 & 0.71 \\
Vanilla RL-Math & Math & 3000 & 40 & 0.43 \\
Vanilla RL-QA & QA & 3000 & 25 & 0.53 \\
Vanilla RL-MMLU & MMLU & 6000 & 94 & 1.0 \\
Vanilla RL-MedQA & Medqa & 3000 & 47 & 1.0 \\
Vanilla RL-MCQ & MMLU\&Medqa & 9000 & 60 & 0.43 \\
\midrule
\multicolumn{5}{c}{\textcolor{gray}{\textbf{7B Models - Ablation Studies}}} \\
Ablation-MCQ-w/o $P_{\text{fallback}}$ & MMLU\&Medqa & 9000 & 110 & 0.78 \\
Ablation-MCQ-w/o Stage-1 & MMLU\&Medqa & 9000 & 110 & 0.78 \\
Ablation-MCQ-w/o Task Mixing & MMLU\&Medqa & 9000 & 120 & 0.85 \\
Ablation-MCQ-QW3 summary & MMLU\&Medqa & 9000 & 50 & 0.36 \\
Ablation-MCQ-4omini summary & MMLU\&Medqa & 9000 & 50 & 0.36 \\
Ablation-Math\&QA weaker Stage-1 & Math\&qa & 6000 & 80 & 0.85 \\
Ablation-QA w/o LLM summary &QA & 3000 & 60 & 1.28 \\
Ablation-QA Raw data &R1-Searcher & 8148 &70 &0.55\\
Ablation-Math Raw data &SimpleRL-Zoo &16662 &10 & 0.056\\
\bottomrule
\end{tabular}
\caption{Training checkpoint details for UR$^2$ models. Checkpoints were saved every 5 steps for 3B models and single-task training, and every 10 steps for larger models. Main experiments use the full training configuration, while ablation studies vary specific components.}
\label{tab:training_checkpoints}
\end{table*}
\subsection{Vanilla RL}
\label{sec:vanilla_rl}
To ensure a fully fair comparison, Vanilla RL is a baseline following the same setup and datasets as UR$^2$, applying RAG-RL to open-domain QA and CoT-RL to math/MCQ tasks. Specifically: (1) identical or subset training data, (2) REINFORCE++ algorithm, (3) same hyperparameters (KL coef, batch size), and (4) standard domain-specific training methods (CoT-RL in reasoning tasks and RAG-RL in open-domain QA). In the retrieval scenario, it serves as a single stage multi-step search–reasoning training baseline similar to Search-R1, using the same setup and subset training data as UR$^2$. In the reasoning scenario, it serves as a single stage onestep reasoning training baseline similar to SimpleRL-Zoo, using the same setup and identical or subset training data as UR$^2$.

\subsection{Evaluation Details}
\label{sec:evaluation_details}

All evaluations are performed using \texttt{vLLM version 0.6.5}. The vLLM version of Qwen-3 used is \texttt{0.8.5.post1}. In evaluation, We maintain the same \texttt{max\_tokens} limits used during training: \texttt{3072} for math benchmarks, \texttt{1536} for MCQ, and \texttt{512} for open-domain QA per generation step. For GPT-family models, these limits are increased to \texttt{4096}, \texttt{2048}, and \texttt{1024}, respectively. For sampling during evaluation, we use more conservative hyperparameters: \texttt{temperature = 0.3} and \texttt{top\_p = 0.5}, aiming for higher answer consistency. Summarization for math tasks is conducted using \texttt{Qwen-3-32B} with \texttt{max\_tokens = 8192}, \texttt{temperature = 0.3}, and \texttt{top\_p = 0.7}. Final evaluation summarization is performed using \texttt{gpt-4.1-2025-04-14} with \texttt{max\_tokens = 2048}, \texttt{temperature = 0.3}, and \texttt{top\_p = 0.5}. 

The RL methods mentioned in this paper all follow the settings described in their original works. Specifically, Open-Reasoner-Zero, General Reasoner, SimpleRL-Zoo, R1-Searcher, Search-R1, and ZeroSearch are implemented using the Qwen-2.5-Base models. Although an Instruct version of Search-R1 exists, its performance is significantly inferior and thus excluded from comparison. R1-Searcher with LLaMA-3.1-8B adopts the Instruct variant. Vanilla methods, including CoT and standard RAG, are applied using the Instruct versions for all open-source models.

\textbf{Advanced RAG Baseline Implementations:}

\textbf{Search-o1 with Retrieval-Augmented Generation}: We adapt the Search-o1 framework~\citep{li2025search} to operate within a controlled evaluation environment. While maintaining its core iterative reasoning mechanism and document analysis capabilities, our implementation leverages the KILT Wikipedia corpus with \texttt{BGE-large-en-v1.5} embeddings for knowledge retrieval. This approach consolidates the multi-agent architecture into a unified model with structured prompting, ensuring consistent evaluation across all baselines while preserving the essential reasoning patterns.

\textbf{Self-Ask with Retrieval-Augmented Generation}: Our implementation follows the Self-Ask framework's~\citep{press2023measuring} question decomposition strategy, employing batch retrieval from the local KILT corpus to enhance efficiency. The system maintains the characteristic ``Follow up:'' and ``Intermediate answer:'' reasoning chain format, with stopping criteria incorporating both semantic completion detection and a maximum of 10 follow-up questions. When decomposition challenges arise, the framework seamlessly transitions to standard RAG, ensuring robust performance across diverse question types.

\textbf{RAT (Retrieval-Augmented Thought)}: We adapt RAT~\citep{wang2024rat} for unified evaluation across reasoning and QA tasks. The framework retains the core principle of knowledge-enhanced reasoning while operating at the paragraph level rather than the sentence level, with corresponding modifications to the prompting strategy. This design choice maintains consistency with our evaluation infrastructure while capturing RAT's fundamental insight of augmenting reasoning processes with relevant external knowledge.

All advanced RAG methods operate within a standardized retrieval infrastructure: documents are retrieved from the 100-word segmented KILT Wikipedia corpus (29M documents in total). For GPT-family models, we use top-$k$=10 retrieval. Due to model limitations, LLaMA and Qwen variants use top-$k$=5. For summarization or other auxiliary operations beyond reasoning, each model performs the processing itself rather than relying on GPT-4.1, ensuring consistency with its own capabilities.

\textbf{Online Corpus Retrieval Implementation:}

To evaluate the generalization capability of UR$^2$ with real-world web content, we implement an online corpus retrieval system that dynamically fetches and processes web documents. Unlike the offline Wikipedia corpus used during training, this online retrieval mechanism provides access to up-to-date information from the internet.

The online retrieval pipeline consists of three main components:

\textbf{Web Search and Content Extraction}: We utilize the \textbf{Bing Search API} to retrieve relevant URLs based on the model's search queries. To ensure robust retrieval quality, we implement a multi-round crawling strategy with up to three rounds of attempts. In each round, the system fetches $k \times 3$ candidate URLs and crawls them in parallel using a thread pool with 256 workers. The system implements intelligent retry logic---if the initial $k$ URLs fail to provide sufficient valid content, it automatically attempts to crawl additional URLs from the candidate pool. This approach significantly improves the success rate of obtaining high-quality content.

\textbf{HTML-to-Markdown Conversion}: Raw HTML content from web pages often contains noise such as navigation elements, advertisements, and scripts. We deploy a dedicated service using ReaderLM-v2-1.5B Model~\footnote{https://huggingface.co/jinaai/reader-lm-1.5b} through the vLLM framework to convert HTML to clean Markdown format. The preprocessing pipeline removes script tags, style elements, base64-encoded images, and other irrelevant content using optimized regular expressions. The model then generates readable Markdown that preserves the main textual information while discarding formatting artifacts. To improve efficiency, we implement an LRU cache with a capacity of 10,000 entries, achieving significant speedup for repeated content.

\textbf{Content Summarization}: The summarization prompt is carefully designed to distinguish between knowledge-based queries (which can be answered with factual information) and reasoning-based queries (which require complex computation). For knowledge-based queries, the model extracts and presents relevant facts; for reasoning-based queries, it returns a fallback message indicating that direct reasoning is more appropriate. The summarizer here is \texttt{GPT-4.1-2025-04-14}.

The entire pipeline is orchestrated through a FastAPI service that handles concurrent requests efficiently. Rate limiting is enforced for the Bing API (95 requests per second) to comply with usage policies. The system maintains detailed logging for debugging and performance monitoring, tracking metrics such as cache hit rates, crawling success rates, and end-to-end latency.

Due to network and hardware limitations, a small portion of Wikipedia pages failed to be crawled correctly, and a subset of queries did not receive valid responses. Given constraints on time and budget, no additional remediation was applied to these cases. However, this reflects the system's alignment with real-world deployment settings, where large-scale QA systems must be robust to occasional retrieval failures and operate under imperfect infrastructure.

This online retrieval implementation enables UR$^2$ to access current information beyond its training data, demonstrating its ability to integrate real-time knowledge into the reasoning process. 
\subsection{Training Dataset Details}
\label{sec:training_dataset_details}
We construct a unified training set that spans multiple task domains to ensure comprehensive coverage of diverse reasoning and knowledge-based challenges. For mathematical reasoning capabilities, we incorporate data from the training split of SimpleZoo-RL, which provides a rich collection of mathematical problem-solving scenarios from \citep{hendrycks2021measuring,cobbe2021training}. Note that since the original SimpleZoo-RL data is relatively simple, medium- and hard-difficulty questions are largely missing, resulting in an overall easy:medium:hard ratio of 1:1:1 rather than the 7:2:1 used in Section~\ref{sec:curriculum_design}. Moreover, due to limitations of LLaMA-3.1-8B-Instruct, we substitute easy-difficulty questions for hard ones during training. To enhance open-domain QA performance, we include samples from the R1-Searcher dataset, which spans a broad range of questions derived from the training sets of 2Wiki and HotpotQA. For specialized domain knowledge, particularly in the medical field, we utilize multi-choice questions from MedQA, ensuring our model can handle domain-specific reasoning in healthcare contexts.

To further diversify our training data and extend coverage to humanities subjects, we generate synthetic questions in three additional domains: philosophy, history, and economics. These synthetic questions are created using \texttt{Qwen-3-32B} and follow the MMLU-Pro format to maintain consistency with established academic evaluation standards. Specifically, we use 5-shot prompting with MMLU-Pro development set examples to generate 10 questions with 4--10 options each. We discard format-non-compliant questions and observe the model's tendency to generate simple questions with 4--5 options, so we request the model to produce additional options and increase the difficulty for each question. For quality control, we use \texttt{GPT-4o-mini-2024-07-18} to evaluate each question's correctness three times, discarding any question identified as incorrect in any evaluation. We then employ Qwen-2.5-7B-Instruct for difficulty assessment, finding approximately 80\% of questions are easy-level. We randomly sample difficult questions as seeds for subsequent generations, using different seeds for each batch. Given that downstream test sets contain subject subdivisions (e.g., Economics encompasses microeconomics, macroeconomics, and econometrics), we utilize \texttt{Qwen-3-32B} to classify questions by subdomain, ensuring comprehensive coverage. We repeat this pipeline for 3--4 iterations to obtain the final training set.

Notably, our synthetic questions differ from MMLU-Pro in emphasizing multi-hop reasoning rather than specific knowledge points. This is evident of our results in Table~\ref{tab:main_reasoning}  where Vanilla RL shows limited improvement over CoT Baseline for Qwen-2.5-7B-Instruct and LLaMA-3.1-8B-Instruct (3.9\% and $-0.1$\% respectively), \textbf{demonstrating no overfitting to the test set}. Despite these characteristics, UR$^2$ consistently achieves improvements across models, validating our method's effectiveness.

\subsection{About Fallback Fault in Retrieval Corpus Construction}
When the policy model generates an invalid search query that triggers a fallback message from the LLM summarizer (i.e., \textit{This query requires design, computation, or complex reasoning, which exceeds the capabilities of a search engine. Please input another query or proceed with direct reasoning.}), we observe that due to the use of retrieval masking, the model gradually learns to treat the content within \texttt{<info>...</info>} as informative for reasoning. As a result, when a fallback fault is encountered, the model tends to hallucinate. Therefore, we append the following visible message after \texttt{</info>} during training to mitigate this issue: \textit{It seems that this query exceeds the capabilities of the retrieval system. We may consider rephrasing it into a more fact-based and searchable question that does not require complex reasoning, or proceed with direct reasoning based on prior knowledge.}

\subsection{Stage 1 Training Details}
\label{sec:stage1_training_details}
Due to the involvement of multiple models and tasks, Section~\ref{sec:stage1} only presents the stage-1 setup for Qwen-2.5-7B-Instruct on math and open-domain QA. Here, we elaborate on the initialization strategies for other models and tasks.

\paragraph{Math and Open-Domain QA.} 
We use the discarded math training samples  with rollout accuracy below 0.2 as cold-start data. These harder examples naturally increase the likelihood of triggering retrieval. For Qwen-2.5-3B-Instruct, its limited capacity makes it more prone to \texttt{Format} violations when invoking retrieval. Since each violation incurs a $-1$ penalty, the original retrieval reward (+3 for one query, +4 for two or more) becomes insufficient to incentivize retrieval. To address this, we increase the retrieval rewards to +5 and +7, respectively. In contrast, LLaMA-3.1-8B-Instruct tends to retrieve for almost every question in early steps. To prevent over-reliance on retrieval and preserve reasoning ability, we remove the extra reward for multiple queries and assign a fixed +3 reward upon any retrieval activation.

\paragraph{MMLU-Pro and Medicine Tasks.}
Unlike math tasks, MMLU-Pro and medicine tasks often require domain-specific knowledge, and retrieval is less likely to lead to fallback faults. For LLaMA-3.1-8B-Instruct and Qwen-2.5-7B-Instruct, a weak reward signal is sufficient during early training: +0.5 for one valid retrieval and +1 for two or more. Unlike the original stage-1 design for math and open-domain QA, this version also incorporates answer rewards from the beginning, facilitating early alignment with task-specific correctness (i.e., no longer relying on cold-start data).In this variant, retrieval rewards are only applied during the first 10 training steps and then disabled. 

For Qwen-2.5-7B-Instruct trained on math and open-domain QA, we adopt the stage-1 setup originally used for the MMLU-Pro and medicine tasks, corresponding to the \texttt{weaker Stage-1} variant in Table~\ref{tab:ablation_qa}. 

For Qwen-2.5-3B-Instruct, we extend Stage 1 to 15 steps. To encourage retrieval, outputs that do not invoke any retrieval call are penalized with a $-1$ \texttt{Format Reward} (non-accumulative).

\subsection{On Randomness and Reproducibility}

RL training is known to exhibit inherent instability and variability across runs, often leading to divergent results even under identical settings~\citep{nagarajan2018deterministic, korkmaz2024understanding}. This randomness is attributed to factors such as stochastic policy updates, environment interactions, and non-deterministic hardware behavior. Despite these challenges, our experiments demonstrate remarkable stability. Thanks to the incorporation of Batch Normalization and Group Normalization in reward calculation, all models converge successfully in a single training run.

During evaluation and result aggregation, we employed a non-zero temperature setting to maintain controlled output diversity, thereby enhancing performance and mitigating the risk of repetitive generations. Due to the substantial API costs associated with GPT-4.1, conducting multiple evaluation runs to average results was not feasible. Nevertheless, given that the datasets contain approximately 500 samples---providing sufficient statistical power---we performed a targeted reproducibility assessment on HotpotQA using the UR$^2$ Qwen 7B-Instruct model. Specifically, three independent evaluation runs yielded F1 scores of 71.7, 71.9, and 71.2, respectively. These consistent results indicate that stochasticity exerts minimal influence on evaluation metrics and comparative model assessments. Furthermore, we conducted supplementary evaluations on all identified outlier cases across baseline and proposed methods to ensure the robustness of our findings.

\subsection{API Consumption}

We measured the API usage cost of UR$^2$ Qwen-2.5-7B-Instruct on MCQ tasks and its \texttt{w/o Stage-1} variant. Training 100 steps with UR$^2$ using GPT-4.1-mini cost approximately \$320, while the \texttt{w/o Stage-1} variant cost around \$100. Additionally, summarization and testing on HotpotQA using GPT-4.1 for UR$^2$ Qwen-2.5-7B-Instruct cost about \$20 per run. Since the training is a one-time expense, we consider the overall training-related consumption acceptable. Furthermore, experiments reported in Section~\ref{sec:corpus_ablation} and Table~\ref{tab:mcqablation} show that substantial performance gains can be achieved without relying on closed-source models, suggesting that open-source models or less expensive APIs provide a viable alternative for achieving comparable improvements.

\subsection{Training Efficiency and Latency}
\label{sec:efficiency}

For models up to 8B parameters with a maximum generation length of 2,048 tokens, UR$^2$ training requires approximately 20 hours of wall-clock time on 8$\times$A100 GPUs, corresponding to ${\sim}$160 GPU-hours in total. All RL baselines are trained under identical hardware and comparable step budgets, so the performance gains are achieved at no additional deployment cost and similar training budgets.

Regarding inference latency, the summarization module introduces an extra generation step per retrieval call. However, by compressing retrieved context from an average of 11,346 tokens to 761 tokens (14.91$\times$), it significantly reduces the reasoning model's input length. Given the quadratic attention cost $O(n^2)$, this compression substantially decreases the downstream reasoning time. In practice, when the summarizer is co-located on the same hardware or served via a fast API, the end-to-end latency of UR$^2$ is comparable to---and in long-context scenarios lower than---standard RAG pipelines that feed raw documents directly. Since UR$^2$ does not increase model parameters, deployment overhead is identical to the base model.

\newpage
\clearpage
\onecolumn
\section{Prompts used in Experiments}
\subsection{Prompts of LLM-as-a-Judge}
\begin{tcolorbox}[colback=gray!5, colframe=black!40, boxrule=0.5pt, title=\textbf{Prompt for Math Evaluation}, breakable, fonttitle=\bfseries, fontupper=\small]
\noindent\makebox[\linewidth]{\hdashrule{\linewidth}{0.5pt}{1mm}}

\textbf{Instruction:}

You are an expert math evaluator.  
Given a question, a gold answer and a predicted answer, judge if they are mathematically consistent.

Ignore formatting (e.g., \texttt{\textbackslash text\{\}}, spacing, capitalization).  
Accept equivalent expressions (e.g., factored vs expanded form).  
If the prediction matches only part of a multi-part answer (e.g., one of several intervals or roots), label it as \textbf{Partially correct}.

\textbf{Output format:}
\begin{itemize}
  \item Reason: Brief explanation  
  \item Judgment: Correct / Partially correct / Incorrect
\end{itemize}

\textbf{Input:}
\begin{itemize}
  \item Question: \texttt{\{question\}}  
  \item Gold: \texttt{\{gold\}}  
  \item Pred: \texttt{\{pred\}}  
\end{itemize}

\noindent\makebox[\linewidth]{\hdashrule{\linewidth}{0.5pt}{1mm}}
\end{tcolorbox}

\begin{tcolorbox}[colback=gray!5, colframe=black!40, boxrule=0.5pt, title=\textbf{Prompt for RAG Evaluation}, breakable, fonttitle=\bfseries, fontupper=\small]
\noindent\makebox[\linewidth]{\hdashrule{\linewidth}{0.5pt}{1mm}}

\textbf{Instruction:}

Given a Question and its Golden Answer, verify whether the Predicted Answer is correct.  
The prediction is correct if it fully aligns with the meaning and key information of the Golden Answer.  
Respond with \texttt{True} if the prediction is correct and \texttt{False} otherwise.

\textbf{Input:}
\begin{itemize}
  \item Question: \texttt{\{question\}}  
  \item Golden Answer: \texttt{\{gold\_answer\}}  
  \item Predicted Answer: \texttt{\{predicted\_answer\}}  
\end{itemize}

\textbf{Your response should be exactly \texttt{"True"} or \texttt{"False"}}

\noindent\makebox[\linewidth]{\hdashrule{\linewidth}{0.5pt}{1mm}}
\end{tcolorbox}
\newpage
\subsection{Prompts of Evaluation and Training}
\label{sec:eval_prompts}
\begin{tcolorbox}[colback=gray!5, colframe=black!40, boxrule=0.5pt, title=\textbf{Prompt for MMLU-Pro\&MedQA}, breakable, fonttitle=\bfseries, fontupper=\small]
\noindent\makebox[\linewidth]{\hdashrule{\linewidth}{0.5pt}{1mm}}

\textbf{Instruction:}

You are solving a multiple-choice question. Analyze each option carefully and logically. Think step by step: consider the meaning and implications of each option, eliminate incorrect ones with clear reasoning, and select the best answer through comparison.

During your reasoning, if you're unsure about any fact, you may issue a \textbf{search query} like this:  
\texttt{<|begin\_of\_query|> your concise query (less than 20 words) <|end\_of\_query|>}

\begin{itemize}
  \item You can issue \textbf{multiple queries} at different steps in your reasoning.
  \item \textbf{Each query must target only one fact or statement}. Do not combine multiple ideas in a single query.
  \item \textbf{Examples:}
    \begin{itemize}
      \item \ding{51} \texttt{<|begin\_of\_query|>} What are the common symptoms of pneumonia? \texttt{<|end\_of\_query|>}
      \item \ding{51}\texttt{<|begin\_of\_query|>} What is the typical treatment for pneumonia in elderly patients? \texttt{<|end\_of\_query|>}
      \item \ding{55}\texttt{<|begin\_of\_query|>} What are the symptoms and treatments for pneumonia in elderly patients? \texttt{<|end\_of\_query|>}
    \end{itemize}
  \item You may issue \textbf{at most four queries} in total --- use them wisely.
\end{itemize}

Once documents are returned in this format:  \\
\texttt{<|begin\_of\_documents|>} ... (search results here) \texttt{<|end\_of\_documents|>}

Use the retrieved documents to verify, reject, or revise your prior reasoning about the options.  
Then continue analyzing the options until you're confident in your answer.

\smallskip
\textbf{Final answer format:}  
\texttt{the correct answer is: A, B, C, D, etc.} (only the letter corresponding to the correct option)

\noindent\makebox[\linewidth]{\hdashrule{\linewidth}{0.5pt}{1mm}}
\end{tcolorbox}

\begin{tcolorbox}[colback=gray!5, colframe=black!40, boxrule=0.5pt, title=\textbf{Prompt for Math}, breakable, fonttitle=\bfseries, fontupper=\small]
\noindent\makebox[\linewidth]{\hdashrule{\linewidth}{0.5pt}{1mm}}

\textbf{Instruction:}

You are solving a math problem. Think step by step to solve it.

The reasoning process includes detailed considerations such as analyzing questions, summarizing relevant findings, brainstorming new ideas, verifying the accuracy of current steps, refining any errors, and revisiting previous steps.

During your reasoning, if you're unsure about a factual concept --- such as a definition, formula, theorem, or mathematical constant --- you may issue a \textbf{search query} to clarify it.

Format your query using the following template (each query must target only one fact):

\texttt{<|begin\_of\_query|>} your concise query (less than 20 words) \texttt{<|end\_of\_query|>}

\textbf{\ding{51} Examples:}
\begin{itemize}
  \item \texttt{<|begin\_of\_query|>} Definition of Möbius function \texttt{<|end\_of\_query|>}
  \item \texttt{<|begin\_of\_query|>} Formula for variance of Bernoulli distribution \\ \texttt{<|end\_of\_query|>}
\end{itemize}

\textbf{\ding{55} Do \underline{NOT} query for reasoning-related content like:}
\begin{itemize}
  \item Whether a solution approach is valid
  \item How to compute a specific value
  \item Multi-step deductions or conclusions
\end{itemize}

You may issue at most \textbf{four} search queries per problem --- use them wisely.

When documents are returned in this format: \\
\texttt{<|begin\_of\_documents|>}  
... (search results here)  
\texttt{<|end\_of\_documents|>}

Use the evidence to confirm or revise your reasoning. Then continue analyzing the question until you're confident in the answer.

At the end of your reasoning, give your final answer in the following format: \\
\texttt{\textbackslash boxed\{YOUR\_ANSWER\}}

\noindent\makebox[\linewidth]{\hdashrule{\linewidth}{0.5pt}{1mm}}
\end{tcolorbox}

\begin{tcolorbox}[colback=gray!5, colframe=black!40, boxrule=0.5pt, title=\textbf{Prompt for Open-Domain QA}, breakable, fonttitle=\bfseries, fontupper=\small]
\noindent\makebox[\linewidth]{\hdashrule{\linewidth}{0.5pt}{1mm}}

\textbf{Instruction:}

You are solving a factual open-domain question from a Knowledge Question Answering (KQA) task. The question requires step-by-step reasoning over real-world knowledge to identify a specific, factually correct answer.

Carefully analyze the question to understand the key entities, relationships, and constraints involved. Retrieve and consider relevant factual knowledge, and reason logically to identify the most accurate answer. 

During your reasoning, if you're unsure about any fact, you may issue a \textbf{search query} like this:  
\texttt{<|begin\_of\_query|>} your concise query (less than 20 words) \texttt{<|end\_of\_query|>}

\begin{itemize}
  \item You can issue \textbf{multiple queries} at different steps in your reasoning.
  \item \textbf{Each query must target only one fact or statement}. Do not combine multiple ideas in a single query.
    \begin{itemize}
      \item \textbf{\ding{51} Example:}
        \begin{itemize}
          \item \texttt{<|begin\_of\_query|>} When did Einstein move to the United States? \texttt{<|end\_of\_query|>}
          \item \texttt{<|begin\_of\_query|>} Why did Einstein leave Germany? \\ \texttt{<|end\_of\_query|>}
        \end{itemize}
      \item \textbf{\ding{55} Do \underline{not} combine them like this:}
        \begin{itemize}
          \item \texttt{<|begin\_of\_query|>} When did Einstein move to the US and why did he leave Germany? \texttt{<|end\_of\_query|>}
        \end{itemize}
    \end{itemize}
  \item You may issue \textbf{at most five queries} in total --- use them wisely.
\end{itemize}

Once documents are returned in this format: \\ 
\texttt{<|begin\_of\_documents|>}  
... (search results here)  
\texttt{<|end\_of\_documents|>}

Use the evidence to confirm or revise your reasoning. Then continue analyzing the question until you're confident in the answer.

At the end of your reasoning, give your final answer in the following format:  
\texttt{\textbackslash boxed\{YOUR\_ANSWER\}}

\noindent\makebox[\linewidth]{\hdashrule{\linewidth}{0.5pt}{1mm}}
\end{tcolorbox}
\newpage
\subsection{Prompts for Summarizing}
\label{sec:summarizing_prompt}
\begin{tcolorbox}[colback=gray!5, colframe=black!40, boxrule=0.5pt, title=\textbf{Prompt for Summarizing Math Documents During Evaluation}, breakable, fonttitle=\bfseries, fontupper=\small]
\noindent\makebox[\linewidth]{\hdashrule{\linewidth}{0.5pt}{1mm}}

\textbf{Task Instruction:}

You are assisting in solving a math problem. You are tasked with reading and analyzing Wikipedia content based on the following inputs: \textbf{Previous Reasoning Steps}, \textbf{Current Search Query}, and \textbf{Wikipedia Content}. Your task is to extract accurate and relevant information from the provided Wikipedia content to support or enhance the reasoning process.

\begin{itemize}
  \item Carefully read the provided \textbf{Wikipedia Content};
  \item Extract factual information that can:
    \begin{itemize}
      \item Directly assist in answering the \textbf{Current Search Query}, or
      \item Help validate, complete, or correct earlier reasoning steps.
    \end{itemize}
  \item The extracted information should be:
    \begin{itemize}
      \item Accurate and trustworthy;
      \item Closely relevant to the query;
      \item Helpful in improving, expanding, or supporting the mathematical reasoning.
    \end{itemize}
\end{itemize}

Important:  
Do NOT attempt to correct or rewrite the previous reasoning. Treat it only as contextual reference that may be flawed.

\textbf{Output Format:}

Present the information beginning with the label \texttt{**Final Information**\\} as shown below.

\texttt{**Final Information**\\}  
[Helpful factual information]

\textbf{Inputs:}
\begin{itemize}
  \item Previous Reasoning Steps: \texttt{\{prev\_reasoning\}}
  \item Current Search Query: \texttt{\{search\_query\}}
  \item Wikipedia Content: \texttt{\{wikipedia\_content\}}
\end{itemize}

\noindent\makebox[\linewidth]{\hdashrule{\linewidth}{0.5pt}{1mm}}
\end{tcolorbox}
\newpage
\begin{tcolorbox}[colback=gray!5, colframe=black!40, boxrule=0.5pt, title=\textbf{Prompt for Summarizing Math Documents During Training}, breakable, fonttitle=\bfseries, fontupper=\small]
\noindent\makebox[\linewidth]{\hdashrule{\linewidth}{0.5pt}{1mm}}

\textbf{Task Instruction:}

You are assisting in solving a math problem. Your task is to determine whether the current query requires external factual knowledge (such as definitions, formulas, theorems, or lookup values), and if so, extract accurate and relevant information from the provided Wikipedia content to support or enhance the reasoning process.

\textbf{Step 1: Classify the Query Type}

Determine whether the query falls into one of the following categories:

\begin{itemize}
  \item \textbf{Knowledge-based query}: Can be directly answered using factual knowledge.
  \item \textbf{Reasoning-based query}: Requires multi-step deduction, logical reasoning, or constructive computation.
\end{itemize}

\textbf{If reasoning-based, return:}
\emph{
This query requires design, computation, or complex reasoning, which exceeds the capabilities of a search engine. Please input another query or proceed with direct reasoning.
}

\textbf{Step 2: Analyze Knowledge-Based Queries (if applicable)}

\begin{itemize}
  \item Carefully read the Wikipedia Content;
  \item Extract factual information that:
    \begin{itemize}
      \item Directly assists the query, or
      \item Helps validate, complete, or correct earlier reasoning.
    \end{itemize}
  \item Ensure information is accurate, relevant, and objective.
\end{itemize}

Do NOT attempt to correct prior reasoning. Treat it as possibly flawed context.

\textbf{Output Format:}

\texttt{**Final Information**\\}  
[Helpful factual information, or the non-knowledge-based response]

\textbf{Inputs:}
\begin{itemize}
  \item Previous Reasoning Steps: \texttt{\{prev\_reasoning\}}
  \item Current Search Query: \texttt{\{search\_query\}}
  \item Wikipedia Content: \texttt{\{wikipedia\_content\}}
\end{itemize}

\noindent\makebox[\linewidth]{\hdashrule{\linewidth}{0.5pt}{1mm}}
\end{tcolorbox}

\begin{tcolorbox}[colback=gray!5, colframe=black!40, boxrule=0.5pt, title=\textbf{Prompt for Summarizing Other Documents During Evaluation}, breakable, fonttitle=\bfseries, fontupper=\small]
\noindent\makebox[\linewidth]{\hdashrule{\linewidth}{0.5pt}{1mm}}

\textbf{Task Instruction:}

You are tasked with reading and analyzing Wikipedia content based on the following inputs: \textbf{Previous Reasoning Steps}, \textbf{Current Search Query}, and \textbf{Wikipedia Content}. Your objective is to extract factual and relevant information from the \textbf{Wikipedia Content} that directly supports or informs the \textbf{Current Search Query}, and integrate it into the reasoning process in an objective and helpful manner.

\textbf{Guidelines:}

\begin{itemize}
  \item Analyze Wikipedia Content:
    \begin{itemize}
      \item Read carefully.
      \item Identify factual info directly related to the query.
    \end{itemize}
  \item Maintain Objectivity:
    \begin{itemize}
      \item Do not validate or revise prior reasoning.
      \item Use it as flawed context.
    \end{itemize}
\end{itemize}

\textbf{Output Format:}

\texttt{**Final Information**\\}  
[Helpful information]

\vspace{+4pt}

\textbf{Inputs:}
\begin{itemize}
  \item Previous Reasoning Steps: \texttt{\{prev\_reasoning\}}
  \item Current Search Query: \texttt{\{search\_query\}}
  \item Wikipedia Content: \texttt{\{wikipedia\_content\}}
\end{itemize}

\noindent\makebox[\linewidth]{\hdashrule{\linewidth}{0.5pt}{1mm}}
\end{tcolorbox}

\begin{tcolorbox}[colback=gray!5, colframe=black!40, boxrule=0.5pt, title=\textbf{Prompt for Summarizing Other Documents During Training}, breakable, fonttitle=\bfseries, fontupper=\small]
\noindent\makebox[\linewidth]{\hdashrule{\linewidth}{0.5pt}{1mm}}

\textbf{Task Instruction:}

Your first task is to determine whether the provided query is a knowledge-based query that can be answered using factual information from Wikipedia, or if it requires design, computation, or complex reasoning.

\textbf{Step 1: Query Classification}
\begin{itemize}
  \item If knowledge-based (e.g., facts, definitions, history), proceed to Step 2.
  \item Otherwise, return:
\end{itemize}
\emph{
This query requires design, computation, or complex reasoning, which exceeds the capabilities of a search engine. Please input another query or proceed with direct reasoning.
}

\textbf{Step 2: Analyze Knowledge-Based Queries}

\begin{itemize}
  \item Read Wikipedia content;
  \item Extract relevant factual information;
  \item Stay neutral---do not alter previous reasoning;
\end{itemize}

\vspace{+4pt}

\textbf{Output Format:}

\texttt{**Final Information**\\}  
[Helpful information or the non-knowledge-based response]

\vspace{+4pt}

\textbf{Inputs:}
\begin{itemize}
  \item Previous Reasoning Steps: \texttt{\{prev\_reasoning\}}
  \item Current Search Query: \texttt{\{search\_query\}}
  \item Wikipedia Content: \texttt{\{wikipedia\_content\}}
\end{itemize}

\noindent\makebox[\linewidth]{\hdashrule{\linewidth}{0.5pt}{1mm}}
\end{tcolorbox}

\newpage

\subsection{Prompts for Baseline Methods}

\begin{tcolorbox}[colback=gray!5, colframe=black!40, boxrule=0.5pt, title=\textbf{Self-Ask Initial Prompt}, breakable, fonttitle=\bfseries, fontupper=\small]
\noindent\makebox[\linewidth]{\hdashrule{\linewidth}{0.5pt}{1mm}}

\textbf{Instruction:}

The self-ask method uses few-shot examples to demonstrate the reasoning pattern:

\textbf{Example 1:}\\
Question: Who lived longer, Muhammad Ali or Alan Turing?\\
Are follow up questions needed here: Yes.\\
Follow up: How old was Muhammad Ali when he died?\\
Intermediate answer: Muhammad Ali was 74 years old when he died.\\
Follow up: How old was Alan Turing when he died?\\
Intermediate answer: Alan Turing was 41 years old when he died.\\
So the final answer is: Muhammad Ali

\textbf{Example 2:}\\
Question: When was the founder of craigslist born?\\
Are follow up questions needed here: Yes.\\
Follow up: Who was the founder of craigslist?\\
Intermediate answer: Craigslist was founded by Craig Newmark.\\
Follow up: When was Craig Newmark born?\\
Intermediate answer: Craig Newmark was born on December 6, 1952.\\
So the final answer is: December 6, 1952

\textbf{Example 3:}\\
Question: Who was the maternal grandfather of George Washington?\\
Are follow up questions needed here: Yes.\\
Follow up: Who was the mother of George Washington?\\
Intermediate answer: The mother of George Washington was Mary Ball Washington.\\
Follow up: Who was the father of Mary Ball Washington?\\
Intermediate answer: The father of Mary Ball Washington was Joseph Ball.\\
So the final answer is: Joseph Ball

\textbf{Input:}\\
Question: \texttt{\{question\}}\\
Options: \texttt{\{options\}}\\
Are follow up questions needed here:

\noindent\makebox[\linewidth]{\hdashrule{\linewidth}{0.5pt}{1mm}}
\end{tcolorbox}

\begin{tcolorbox}[colback=gray!5, colframe=black!40, boxrule=0.5pt, title=\textbf{Self-Ask Sub-question Answering Prompt}, breakable, fonttitle=\bfseries, fontupper=\small]
\noindent\makebox[\linewidth]{\hdashrule{\linewidth}{0.5pt}{1mm}}

\textbf{Instruction:}

Please answer the following question based on the reference text. If the reference text does not contain sufficient information to answer the question, you may use your own knowledge to provide the answer. Always think step by step.

Provide your final answer in the format \texttt{\textbackslash boxed\{YOUR\_ANSWER\}}.

\textbf{Input:}
\begin{itemize}
  \item Question: \texttt{\{subquestion\}}
  \item Reference text: \texttt{\{reference\}}
\end{itemize}

\noindent\makebox[\linewidth]{\hdashrule{\linewidth}{0.5pt}{1mm}}
\end{tcolorbox}

\newpage

\begin{tcolorbox}[colback=gray!5, colframe=black!40, boxrule=0.5pt, title=\textbf{RAT Draft Generation Prompt}, breakable, fonttitle=\bfseries, fontupper=\small]
\noindent\makebox[\linewidth]{\hdashrule{\linewidth}{0.5pt}{1mm}}

\textbf{System Prompt:}

You are an advanced AI assistant tasked with answering open-domain questions. You excel at providing comprehensive, well-structured answers with multiple paragraphs. Each paragraph you write contains multiple sentences that thoroughly explore the topic. You always follow formatting instructions precisely.

\textbf{Instruction:}

IMPORTANT: Structure your response as follows:

1. Write a comprehensive answer with MULTIPLE PARAGRAPHS (3-6 paragraphs typically).

2. Each paragraph MUST contain AT LEAST 2 complete sentences. Single-sentence paragraphs are NOT acceptable.

3. Separate paragraphs with blank lines (press Enter twice).

4. At the very end, after all paragraphs, add your final answer in this format:\\
   \texttt{\textbackslash box\{ANSWER\}}
   
   where ANSWER is ONLY the direct answer - typically just a name, number, date, or short phrase.
   Examples:
   \begin{itemize}
     \item For ``Who was the first president?'' → \texttt{\textbackslash box\{George Washington\}}
     \item For ``When was the company founded?'' → \texttt{\textbackslash box\{1812\}}
     \item For ``What is the capital?'' → \texttt{\textbackslash box\{Paris\}}
   \end{itemize}
   
   DO NOT include explanations or full sentences in the box.

\textbf{Input:}
\begin{itemize}
  \item Question: \texttt{\{question\}}
\end{itemize}

\noindent\makebox[\linewidth]{\hdashrule{\linewidth}{0.5pt}{1mm}}
\end{tcolorbox}

\begin{tcolorbox}[colback=gray!5, colframe=black!40, boxrule=0.5pt, title=\textbf{RAT Query Generation Prompt}, breakable, fonttitle=\bfseries, fontupper=\small]
\noindent\makebox[\linewidth]{\hdashrule{\linewidth}{0.5pt}{1mm}}

\textbf{Instruction:}

Based on the question and the current answer content, generate a search query to verify or find additional information.

Please summarize the content with the corresponding question. This summarization will be used as a query to search with Bing search engine. The query should be short but need to be specific to promise Bing can find related knowledge or pages. You can also use search syntax to make the query short and clear enough for the search engine to find relevant language data. Try to make the query as relevant as possible to the last few sentences in the content.

\textbf{IMPORTANT:} Just output the query directly. DO NOT add additional explanations or introducement in the answer unless you are asked to.

\textbf{Input:}
\begin{itemize}
  \item Question: \texttt{\{question\}}
  \item Current Answer: \texttt{\{current\_answer\}}
\end{itemize}

\noindent\makebox[\linewidth]{\hdashrule{\linewidth}{0.5pt}{1mm}}
\end{tcolorbox}

\newpage

\begin{tcolorbox}[colback=gray!5, colframe=black!40, boxrule=0.5pt, title=\textbf{RAT Answer Revision Prompt}, breakable, fonttitle=\bfseries, fontupper=\small]
\noindent\makebox[\linewidth]{\hdashrule{\linewidth}{0.5pt}{1mm}}

\textbf{Instruction:}

I want to revise the answer according to retrieved related text of the question. You need to check whether the answer is correct. If you find some errors in the answer, revise the answer to make it better. If you find some necessary details are ignored, add it to make the answer more plausible according to the related text.

\textbf{IMPORTANT:}
\begin{enumerate}
  \item Keep the structure with multiple substantial paragraphs.
  \item Use blank lines to separate paragraphs (press Enter twice).
  \item If the original answer has \texttt{\textbackslash box\{...\}} at the end, you MUST keep it and update it if needed.
  \item The \texttt{\textbackslash box\{\}} should contain ONLY the direct answer (name/number/date/short phrase), NOT a full sentence.
\end{enumerate}

Just output the revised paragraphs directly, including the \texttt{\textbackslash box\{\}} if present.

\textbf{Input:}
\begin{itemize}
  \item Retrieved Text: \texttt{\{retrieved\_text\}}
  \item Question: \texttt{\{question\}}
  \item Answer: \texttt{\{current\_answer\}}
\end{itemize}

\noindent\makebox[\linewidth]{\hdashrule{\linewidth}{0.5pt}{1mm}}
\end{tcolorbox}

\begin{tcolorbox}[colback=gray!5, colframe=black!40, boxrule=0.5pt, title=\textbf{Search-o1 Reasoning Prompt}, breakable, fonttitle=\bfseries, fontupper=\small]
\noindent\makebox[\linewidth]{\hdashrule{\linewidth}{0.5pt}{1mm}}

\textbf{System Prompt:}

You are a reasoning assistant with the ability to perform web searches to help you answer the user's question accurately. You have special tools:

\begin{itemize}
  \item To perform a search: write \texttt{<|begin\_search\_query|>} your query here \texttt{<|end\_search\_query|>}.
  \item Then, the system will search and analyze relevant web pages, then provide you with helpful information in the format \texttt{<|begin\_search\_result|>} ...search results... \texttt{<|end\_search\_result|>}.
\end{itemize}

You can repeat the search process multiple times if necessary. The maximum number of search attempts is limited to \texttt{\{max\_rounds\}}.

Once you have all the information you need, continue your reasoning.

\textbf{Example:}\\
Question: ``Alice David is the voice of Lara Croft in a video game developed by which company?''

Assistant thinking steps:
\begin{itemize}
  \item I need to find out who voices Lara Croft in the video game.
  \item Then, I need to determine which company developed that video game.
\end{itemize}
\end{tcolorbox}

\newpage

\begin{tcolorbox}[colback=gray!5, colframe=black!40, boxrule=0.5pt, title=\textbf{Prompt for MMLU-Pro\&MedQA (CoT)}, breakable, fonttitle=\bfseries, fontupper=\small]
\noindent\makebox[\linewidth]{\hdashrule{\linewidth}{0.5pt}{1mm}}

\textbf{Instruction:}

You are solving a multiple-choice question. Think step by step and use careful reasoning.
For each question, \textbf{analyze all options one by one}. For each option:
\begin{itemize}
  \item Consider its meaning and implications.
  \item Evaluate whether it is correct or incorrect, and \textbf{explain why}.
  \item Eliminate incorrect options with clear, logical reasoning.
\end{itemize}
After analyzing all options, compare the remaining ones and choose the best answer.

At the end of your reasoning, give your final answer in the following format:\\
\texttt{the correct answer is: A, B, C, D, etc.} (only the letter corresponding to the correct option).

\textbf{Input:}
\begin{itemize}
  \item Question: \texttt{\{question\}}
  \item Options: \texttt{\{options\}}
\end{itemize}

\noindent\makebox[\linewidth]{\hdashrule{\linewidth}{0.5pt}{1mm}}
\end{tcolorbox}

\begin{tcolorbox}[colback=gray!5, colframe=black!40, boxrule=0.5pt, title=\textbf{Prompt for Math (CoT)}, breakable, fonttitle=\bfseries, fontupper=\small]
\noindent\makebox[\linewidth]{\hdashrule{\linewidth}{0.5pt}{1mm}}

\textbf{Instruction:}

Please answer the following math question. You should think step by step to solve it.

Provide your final answer in the format \texttt{\textbackslash boxed\{YOUR\_ANSWER\}}.

\textbf{Input:}
\begin{itemize}
  \item Question: \texttt{\{question\}}
\end{itemize}

\noindent\makebox[\linewidth]{\hdashrule{\linewidth}{0.5pt}{1mm}}
\end{tcolorbox}

\begin{tcolorbox}[colback=gray!5, colframe=black!40, boxrule=0.5pt, title=\textbf{Prompt for Open-Domain QA (CoT)}, breakable, fonttitle=\bfseries, fontupper=\small]
\noindent\makebox[\linewidth]{\hdashrule{\linewidth}{0.5pt}{1mm}}

\textbf{Instruction:}

\texttt{\{question\}}

Please reason step by step, and put your final answer within \texttt{\textbackslash boxed\{\}}.

\noindent\makebox[\linewidth]{\hdashrule{\linewidth}{0.5pt}{1mm}}
\end{tcolorbox}

\tcbset{
  mycasefirst/.style={
    enhanced,
    colframe=black!100,
    boxrule=0.1pt,
    fonttitle=\bfseries,
    fontupper=\footnotesize,
    breakable,
    colback=white,
    overlay={
      \draw[black!100, line width=0.1pt] (frame.north west) rectangle (frame.south east);
      \fill[myblue!100] (frame.north west) rectangle ([yshift=-4cm]frame.north east);

    }
  }
}
\tcbset{
  mycasesecond/.style={
    enhanced,
    colframe=black!100,
    boxrule=0.1pt,
    fonttitle=\bfseries,
    fontupper=\footnotesize,
    colback=white,
    overlay={
      \draw[black!100, line width=0.1pt] (frame.north west) rectangle (frame.south east);
      \fill[myblue!100] (frame.north west) rectangle ([yshift=-8cm]frame.north east);

    }
  }
}
\tcbset{
  mycasethird/.style={
    enhanced,
    colframe=black!100,
    boxrule=0.1pt,
    fonttitle=\bfseries,
    fontupper=\footnotesize,
    colback=white,
    overlay={
      \draw[black!100, line width=0.1pt] (frame.north west) rectangle (frame.south east);
      \fill[myblue!100] (frame.north west) rectangle ([yshift=-4cm]frame.north east);

    }
  }
}
\tcbset{
  mycasefourth/.style={
    enhanced,
    colframe=black!100,
    boxrule=0.1pt,
    fonttitle=\bfseries,
    fontupper=\footnotesize,
    colback=white,
    overlay={
      \draw[black!100, line width=0.1pt] (frame.north west) rectangle (frame.south east);
      \fill[myblue!100] (frame.north west) rectangle ([yshift=-8.2cm]frame.north east);

    }
  }
}
\tcbset{
  mycaseplain/.style={
    enhanced,
    colframe=black!100,
    boxrule=0.1pt,
    fonttitle=\bfseries,
    fontupper=\footnotesize,
    colback=white,
    arc=8pt,  
    outer arc=8pt
  }
}

\newtcolorbox{chatbubbleredbone}[2][]{%
  enhanced,
  sharp corners,
  colback=red!10,      
  colframe=red,        
  boxrule=0.8pt,
  width=2.5cm,
  fontupper=\small,
  #1
}
\newtcolorbox{chatbubblegreenbone}[2][]{%
  enhanced,
  sharp corners,
  colback=mydarkgreen!10,      
  colframe=mydarkgreen,        
  boxrule=0.8pt,
  width=2cm,
  fontupper=\small,
  overlay={
    \draw[mydarkgreen, thick] ([xshift=10mm,yshift=0mm]frame.north west) -- ++(-9mm,9mm); 

  },
  #1
}
\newtcolorbox{chatbubblegreenbtwoleft}[2][]{%
  enhanced,
  sharp corners,
  colback=mydarkgreen!10,
  colframe=mydarkgreen,
  boxrule=0.8pt,
  width=3cm,
  fontupper=\small,
  overlay={
    \draw[mydarkgreen, thick] ([xshift=10mm,yshift=0mm]frame.north west) -- ++(-9mm,9mm); },
  #1
}
\newtcolorbox{chatbubbleyellowbtwoleft}[2][]{%
  enhanced,
  sharp corners,
  colback=mydarkyellow!10,
  colframe=mydarkyellow,
  boxrule=0.8pt,
  width=2cm,
  fontupper=\small,
  overlay={
    \draw[mydarkyellow, thick] ([xshift=10mm,yshift=0mm]frame.north west) -- ++(-3mm,3mm); },
  #1
}
\newtcolorbox{chatbubbleredbtwoleft}[2][]{%
  enhanced,
  sharp corners,
  colback=red!10,
  colframe=red,
  boxrule=0.8pt,
  width=2cm,
  fontupper=\small,
  overlay={
    \draw[red, thick] ([xshift=6mm,yshift=0mm]frame.north west) -- ++(-9mm,9mm); },
  #1
}
\newtcolorbox{chatbubblegreenbtworight}[2][]{%
  enhanced,
  sharp corners,
  colback=mydarkgreen!10,
  colframe=mydarkgreen,
  boxrule=0.8pt,
  width=3.7cm,
  fontupper=\small,
  overlay={
    \draw[mydarkgreen, thick] ([xshift=18mm,yshift=0mm]frame.north west) -- ++(-3mm,3mm); },
  #1
}
\newtcolorbox{chatbubbleyellowbtworight}[2][]{%
  enhanced,
  sharp corners,
  colback=mydarkyellow!10,
  colframe=mydarkyellow,
  boxrule=0.8pt,
  width=3.7cm,
  fontupper=\small,
  overlay={
    \draw[mydarkyellow, thick] ([xshift=15mm,yshift=0mm]frame.north west) -- ++(-3mm,3mm); },
  #1
}
\newtcolorbox{chatbubbleredbtworight}[2][]{%
  enhanced,
  sharp corners,
  colback=red!10,
  colframe=red,
  boxrule=0.8pt,
  width=3.7cm,
  fontupper=\small,
  overlay={
    \draw[red, thick] ([xshift=6mm,yshift=0mm]frame.south west) -- ++(-2mm,-2mm); },
  #1
}
\newtcolorbox{chatbubbleyellowbthreeleft}[2][]{%
  enhanced,
  sharp corners,
  colback=mydarkyellow!10,
  colframe=mydarkyellow,
  boxrule=0.8pt,
  width=3.8cm,
  fontupper=\small,
  overlay={
    \draw[mydarkyellow, thick] ([xshift=15mm,yshift=0mm]frame.north west) -- ++(-3mm,3mm); },
  #1
}
\newtcolorbox{chatbubbleyellowbfour}[2][]{%
  enhanced,
  sharp corners,
  colback=mydarkyellow!10,
  colframe=mydarkyellow,
  boxrule=0.8pt,
  width=3cm,
  fontupper=\small,
  overlay={
    \draw[mydarkyellow, thick] ([xshift=0mm,yshift=-5mm]frame.north west) -- ++(-30mm,8mm); },
  #1
}
\newtcolorbox{chatbubbleyellowbfive}[2][]{%
  enhanced,
  sharp corners,
  colback=mydarkyellow!10,
  colframe=mydarkyellow,
  boxrule=0.8pt,
  width=3cm,
  fontupper=\small,
  overlay={
    \draw[mydarkyellow, thick] ([xshift=8mm,yshift=0mm]frame.north west) -- ++(-3mm,3mm); },
  #1
}
\newtcolorbox{chatbubbleyellowbsix}[2][]{%
  enhanced,
  sharp corners,
  colback=mydarkyellow!10,
  colframe=mydarkyellow,
  boxrule=0.8pt,
  width=4cm,
  fontupper=\small,
  overlay={
    \draw[mydarkyellow, thick] ([xshift=8mm,yshift=0mm]frame.north west) -- ++(-3mm,3mm); },
  #1
}
\newpage
\section{Case Study}
\label{case_study}
\subsection{Search independently based on different difficulty levels and questions}
\begin{figure*}[htbp!]
\centering
    \includegraphics[width=\columnwidth]{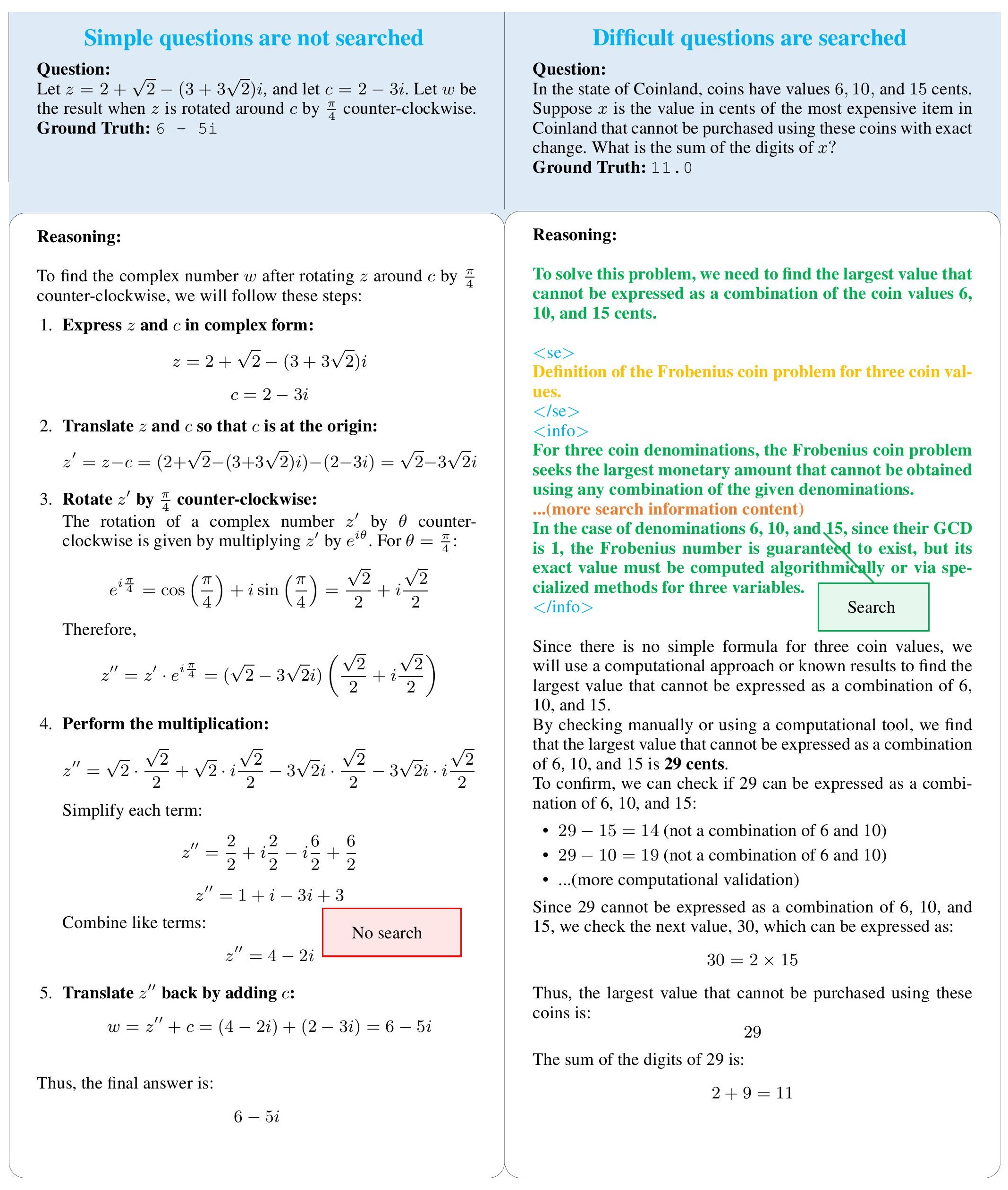}
\caption{UR$^2$ can search independently based on different difficulty levels and questions. \texttt{\bfseries Left:}For simpler questions, UR$^2$ considers that external knowledge support is not necessary and does not perform a search. \texttt{\bfseries Right:}For questions of a certain difficulty, UR$^2$ will actively search for answers.}
\label{fig:case1}
\end{figure*}

\newpage

\subsection{Planning}
\begin{figure*}[htbp!]
\centering
    \includegraphics[width=\columnwidth]{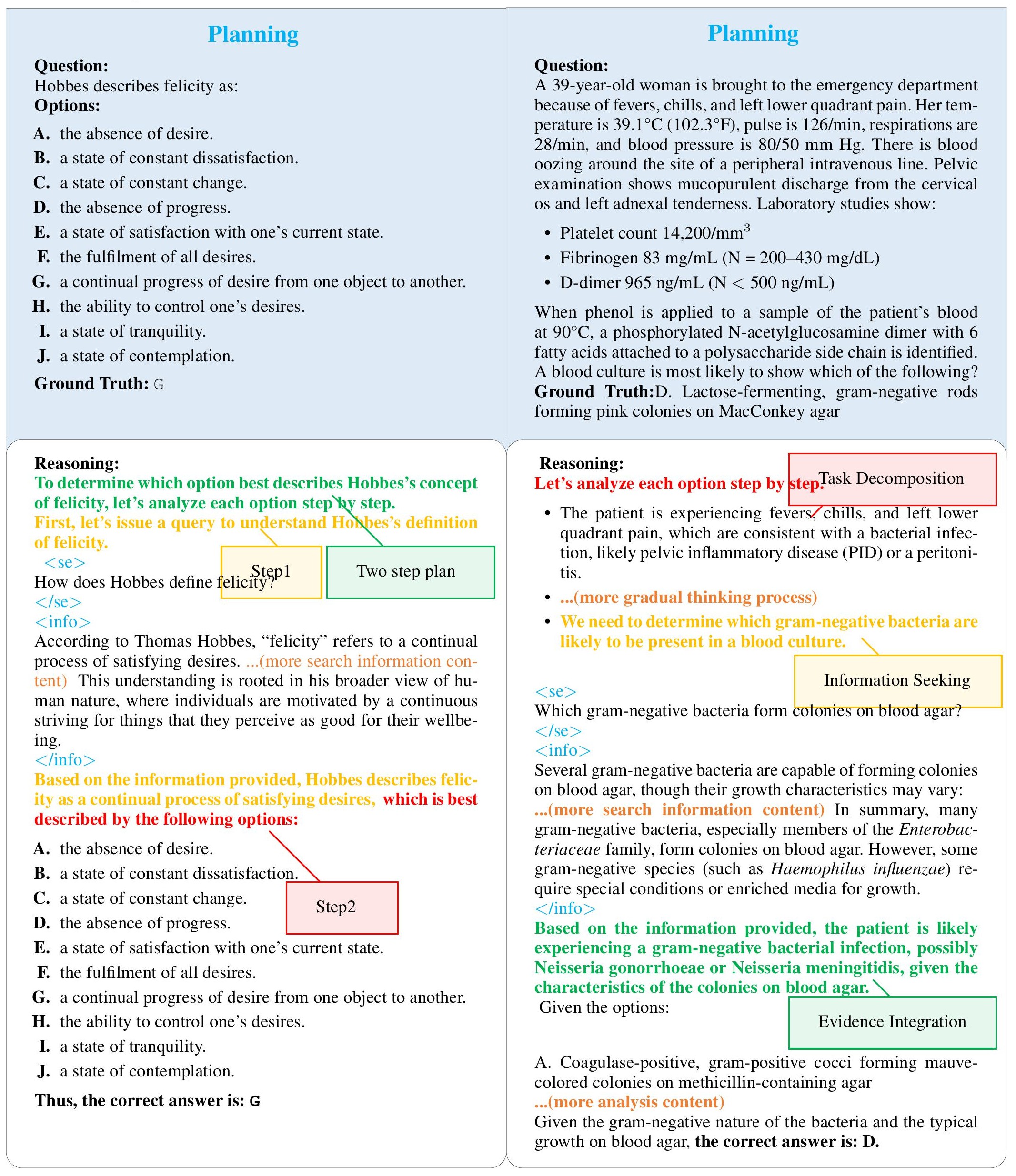}
\caption{UR$^2$ can formulate plans and dynamically adjust them during the reasoning process. \texttt{\bfseries Left:}UR$^2$ develops a plan and completes it in two steps. The first step is to search for the required knowledge, and the second step is to check each option individually. \texttt{\bfseries Right:}UR$^2$ demonstrates clear planning behavior by decomposing the diagnostic task into sequential reasoning steps, identifying knowledge gaps, and querying external information to support its final decision.}
\label{fig:case3}
\end{figure*}
 
\newpage

\subsection{Cross Validation}
\begin{figure*}[ht!]
\centering
    \includegraphics[width=\columnwidth]{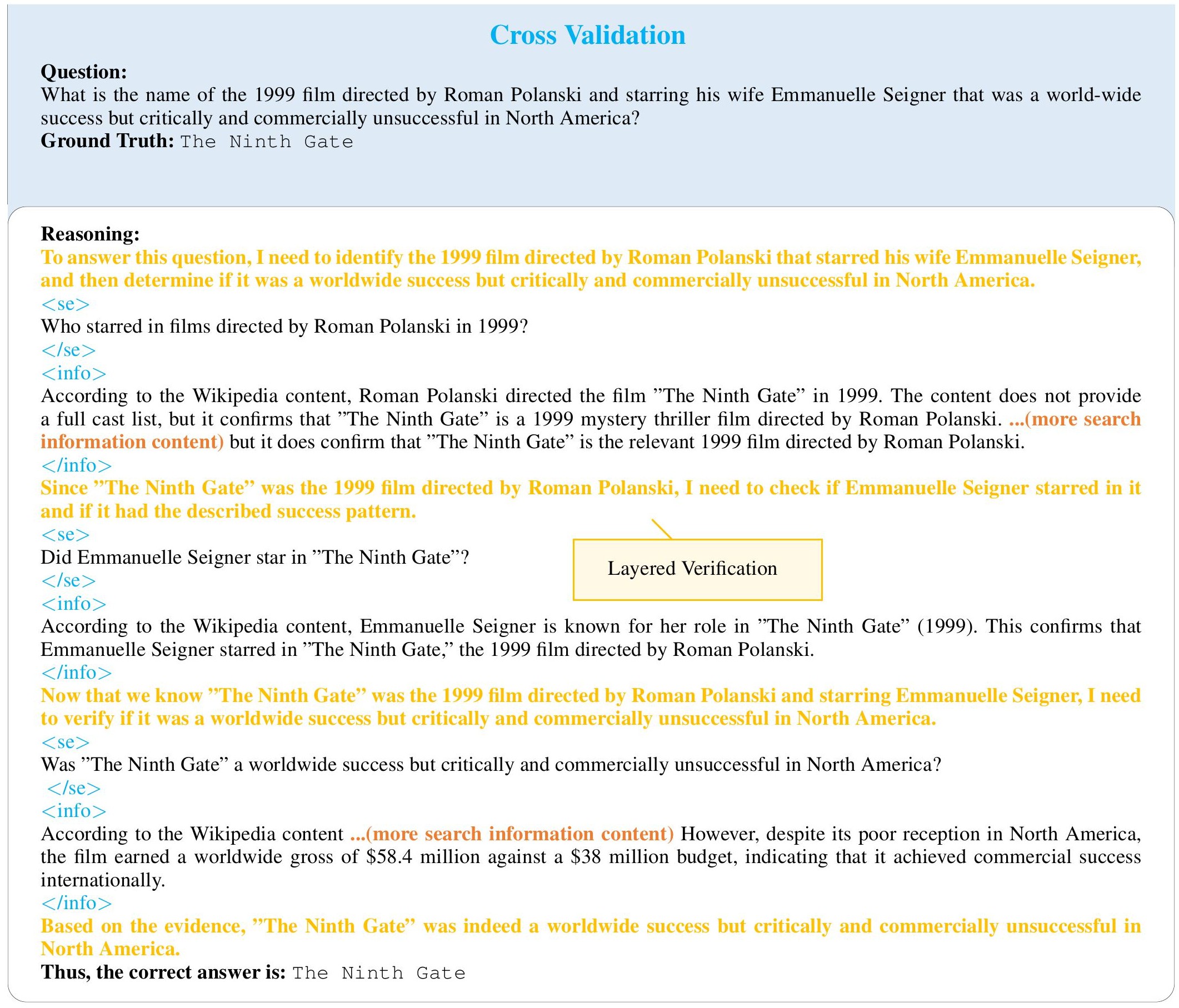}
\caption{This case demonstrates UR$^2$ cross validation ability by verifying the same candidate answer "The Ninth Gate" from multiple angles. It confirms the director and year, validates the actress's participation, and contrasts international vs. North American commercial and critical success. The multi-step confirmation strengthens the confidence in the final answer, showcasing robust cross validation behavior.}
\end{figure*}

\newpage

\subsection{Reflection(Retry Retrieve/Refine Query)}

\begin{figure*}[ht!]
    \includegraphics[width=\columnwidth]{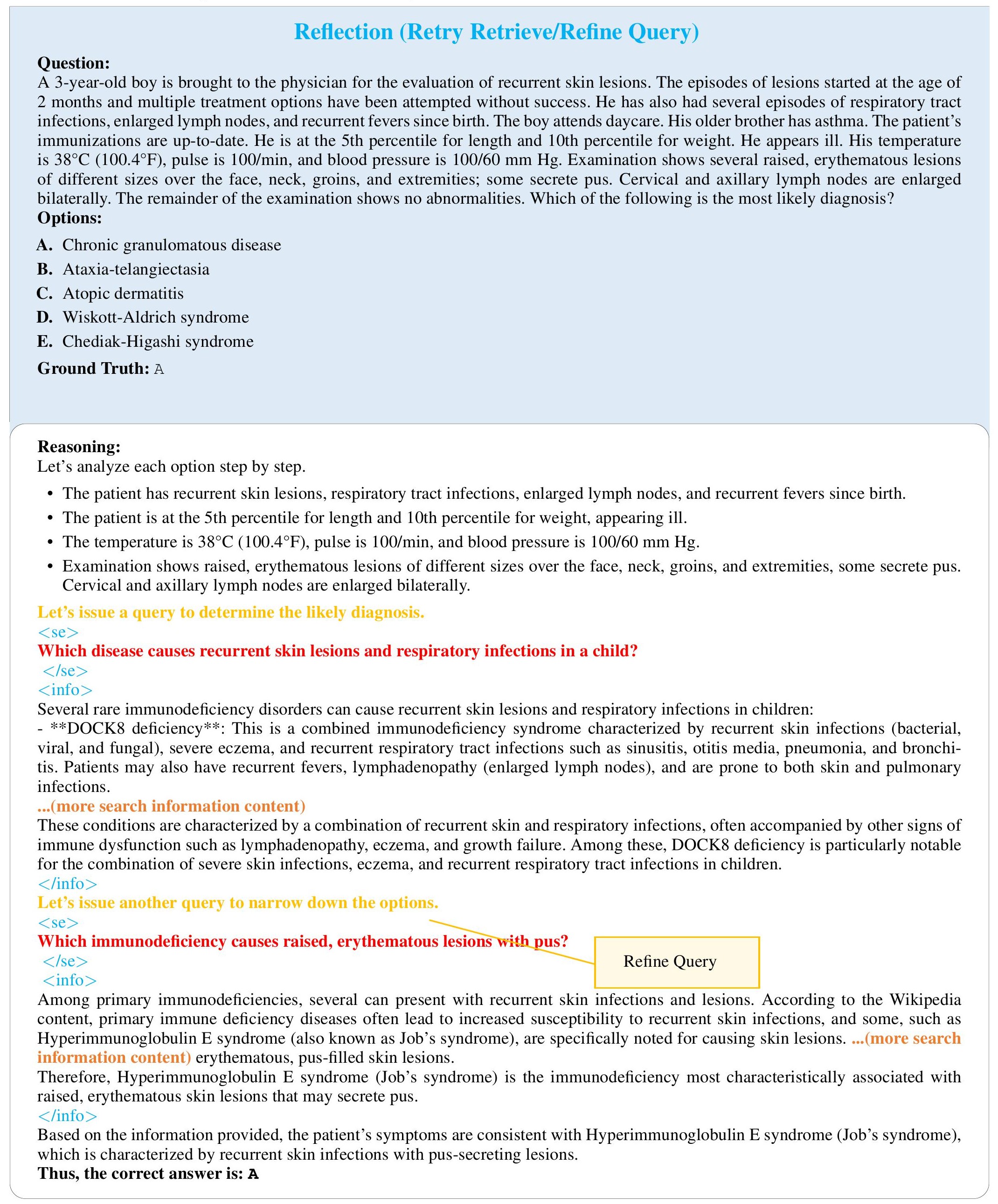}
\caption{After an initial analysis of the patient's symptoms,UR² conducted two rounds of retrieval: first a broad query ("Which disease causes..."), then refined to a targeted question ("Which immunodeficiency causes...") after recognizing the initial query was insufficient. This fully demonstrates the LLM's capability to refine queries.
}
\label{fig:case4}
\end{figure*}

\begin{figure*}[ht!]
    \includegraphics[width=\columnwidth]{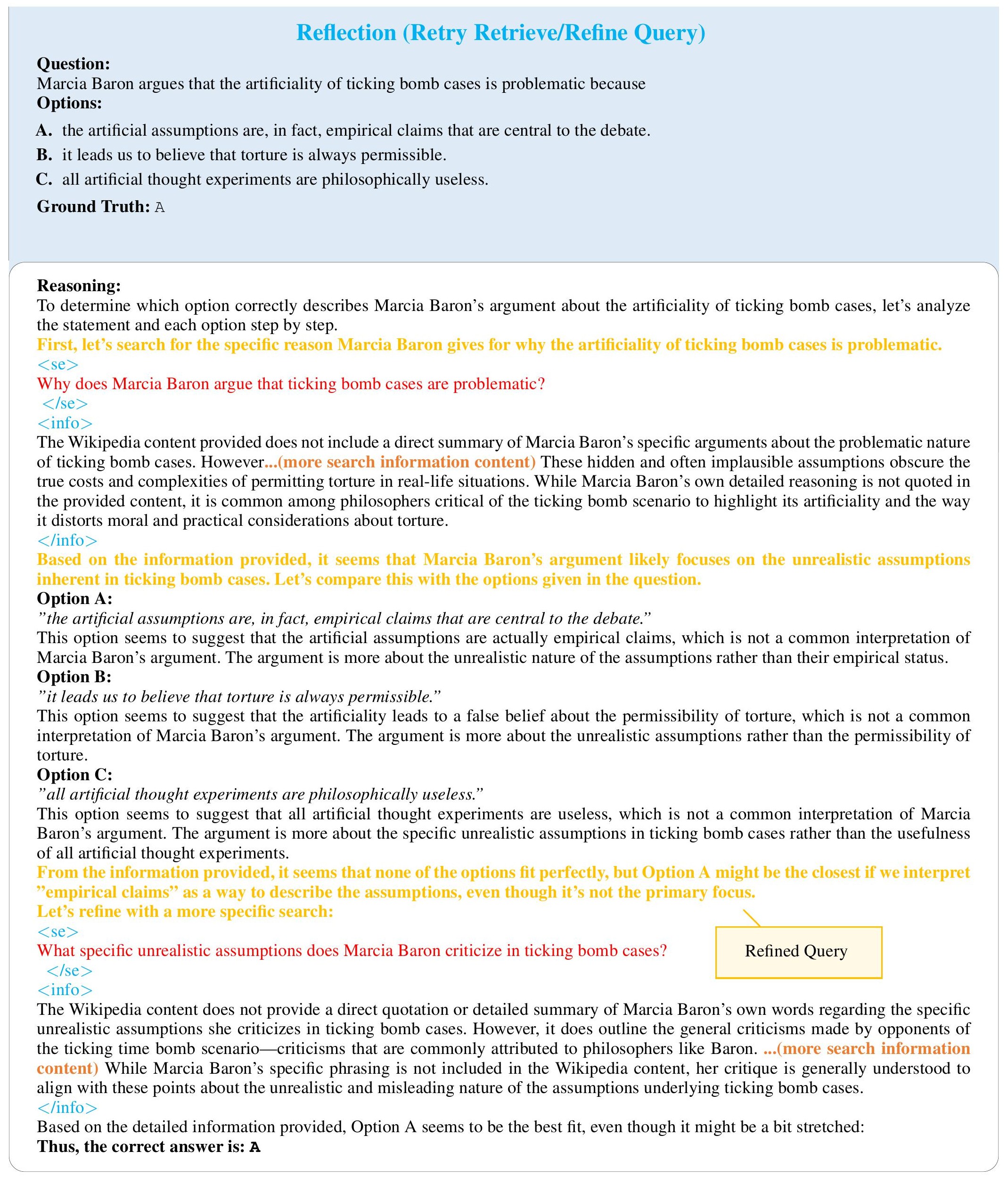}
\caption{In this case, UR$^2$ initially retrieved general critiques related to ticking bomb scenarios but found no direct reference to Marcia Baron's argument. Recognizing this gap, it engaged in Reflection by refining the query to focus on ``specific unrealistic assumptions'', thereby exemplifying its capability to self-correct and deepen its information retrieval process.}
\label{fig:case5}
\end{figure*}

\begin{figure*}[ht!]
    \includegraphics[width=\columnwidth]{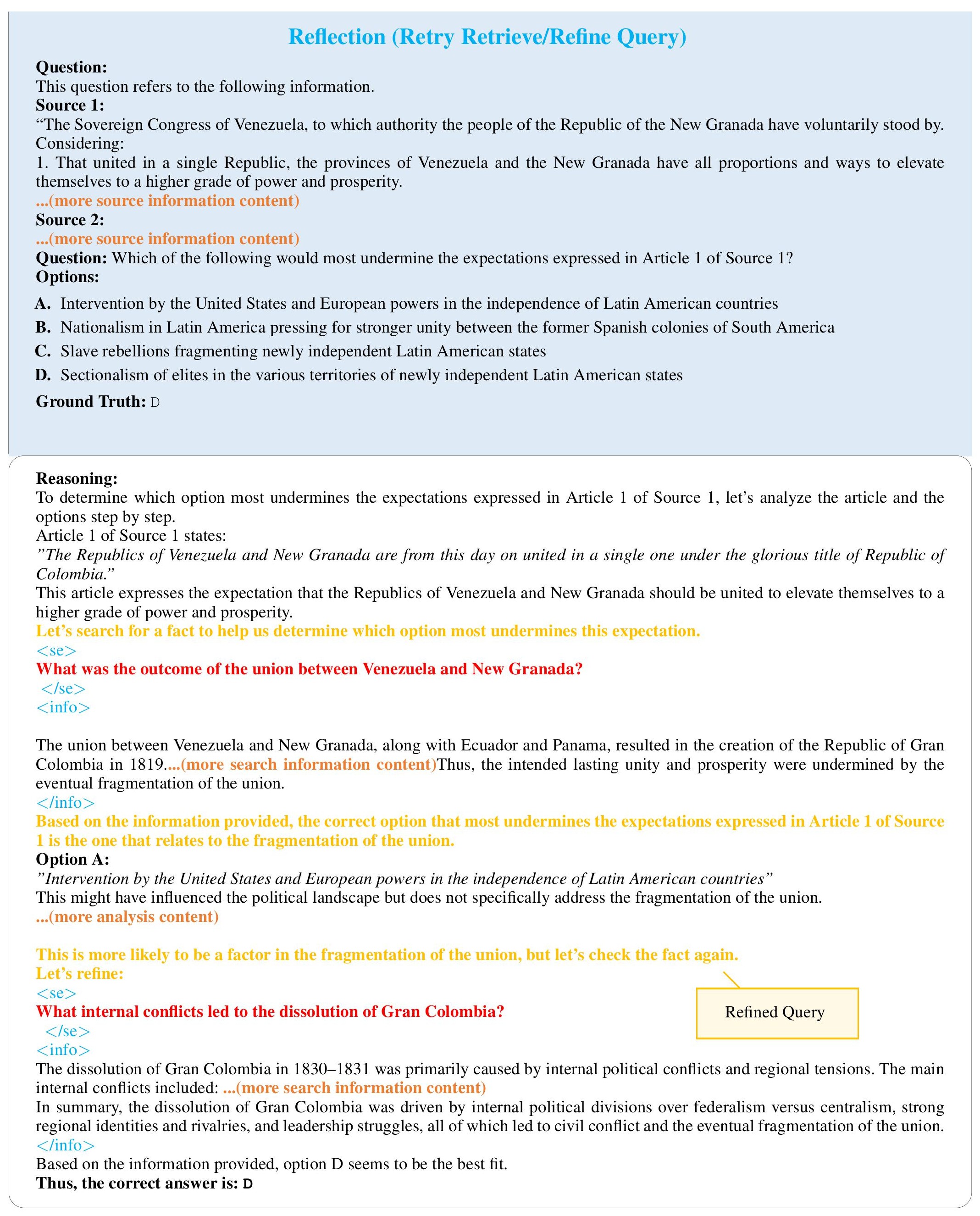}
\caption{UR$^2$ initially retrieved general information about Gran Colombia's dissolution, but upon reflection recognized this was insufficient for precise answer selection. It then refined the query to investigate specific internal conflicts, ultimately identifying elite sectionalism as the key undermining factor, demonstrating effective use of Reflection through query refinement.}
\label{fig:case6}

\end{figure*}

\begin{figure*}[ht!]
    \includegraphics[width=\columnwidth]{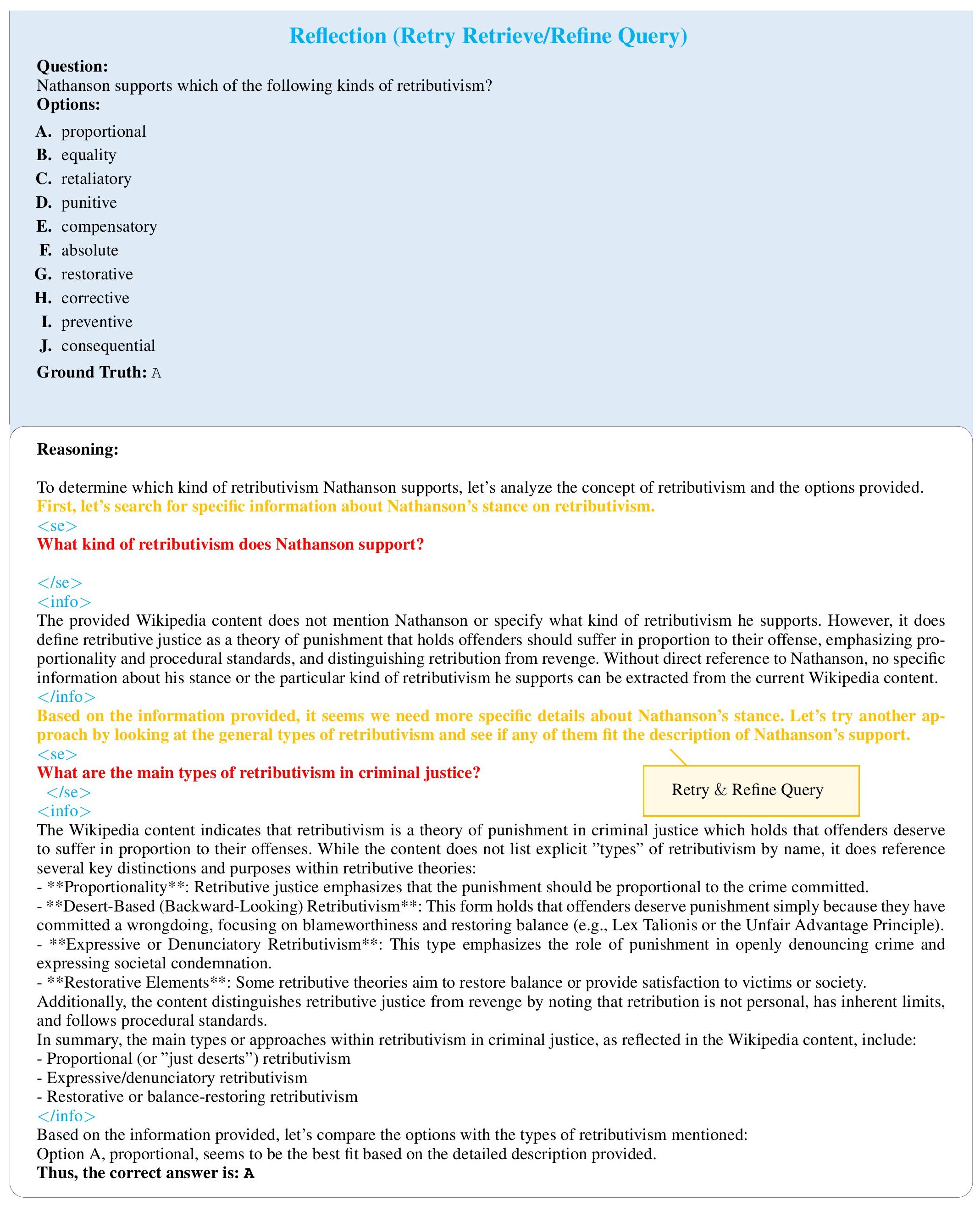}
\caption{This case demonstrates UR$^2$ capacity to reflect and refine its query strategy when initial retrieval fails. By shifting from a direct search for Nathanson's view to analyzing types of retributivism, the model successfully deduced the answer through indirect reasoning. This exemplifies its performance in reflection and retry-based retrieval refinement.}
\label{fig:case7}
\end{figure*}

\end{document}